\documentclass[11pt]{article}
\usepackage{enumitem}
\usepackage{xcolor}
\definecolor{msftBlack}{RGB}{0,0,0}

\usepackage[final]{acl}

\usepackage{times}
\usepackage{latexsym}
\usepackage{amsmath}
\usepackage{booktabs}
\usepackage{multirow} 
\usepackage[table]{xcolor}  
\usepackage{enumitem} 
\usepackage{booktabs}      
\usepackage{listings} 
\usepackage{graphicx}      
\usepackage[table]{xcolor} 
\usepackage{makecell}      
\usepackage{amssymb}       
\usepackage{subcaption}
\usepackage{tikz}
\newcommand*\circled[1]{\tikz[baseline=(char.base)]{
            \node[shape=circle,draw,inner sep=1pt] (char) {#1};}}

\lstdefinestyle{appendixprompt}{
  basicstyle=\ttfamily\footnotesize,
  columns=fullflexible,
  breaklines=true,
  breakatwhitespace=true,
  keepspaces=true,
  showstringspaces=false,
  frame=single,
  framerule=0.3pt,
  rulecolor=\color{black!25},
  xleftmargin=0.6em,
  xrightmargin=0.2em,
  aboveskip=0.6em,
  belowskip=0.6em
}

\usepackage{pifont}
\newcommand{\cmark}{\ding{51}} 
\newcommand{\xmark}{\ding{55}} 

\usepackage[T1]{fontenc}

\usepackage[utf8]{inputenc}

\usepackage{microtype}

\usepackage{inconsolata}

\usepackage{graphicx}

\usepackage[tikz]{bclogo}
\newcommand{\finding}[1]{
\begin{bclogo}[couleur= msftBlack!05, epBord= 1, arrondi=0.1, logo=\bclampe,marge= 2, ombre=true, blur, couleurBord=msftBlack!10, tailleOndu=3, sousTitre ={\em #1}]{} 
\end{bclogo}
}

\definecolor{forestgreen}{HTML}{3B7D23} 
\definecolor{burntorange}{HTML}{C04F15} 

\title{%
\raisebox{-0.2\height}{\includegraphics[height=1.2em]{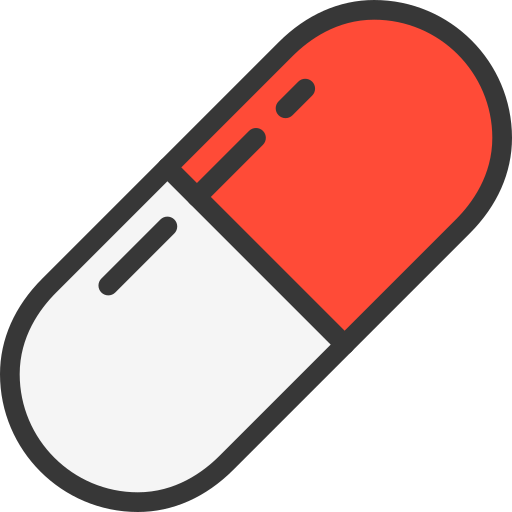}}%
\hspace{0.4em}%
MEDSYN: Benchmarking Multi-EviDence SYNthesis in Complex Clinical Cases for Multimodal Large Language Models}

\author{
 \textbf{Boqi Chen\textsuperscript{1,2,}\footnotemark[2]},
 \textbf{Xudong Liu\textsuperscript{3,}\footnotemark[2]\textsuperscript{,}\thanks{Work done before joining Amazon.}},
 \textbf{Jiachuan Peng\textsuperscript{2,4}},
 \textbf{Marianne Frey-Marti\textsuperscript{5}},
\\
 \textbf{Kyle Lam\textsuperscript{6}},
 \textbf{Bang Zheng\textsuperscript{7}},
 \textbf{Lin Li\textsuperscript{4}},
 \textbf{Jianing Qiu\textsuperscript{2}}
\\
\\
 \textsuperscript{1}ETH Zurich,
 \textsuperscript{2}MBZUAI,
 \textsuperscript{3}Amazon,
 \textsuperscript{4}University of Oxford,
\\
 \textsuperscript{5}University of Bern,
 \textsuperscript{6}Imperial College London
 \textsuperscript{7}Peking University,
\\
 \small{
   \footnotemark[2] Equal contribution.\quad
   \textbf{Correspondence:} \href{mailto:email@domain}{Jianing.Qiu@mbzuai.ac.ae}
 }
}

\begin{document}
\maketitle
\begin{abstract}
Multimodal large language models (MLLMs) have shown great potential in medical applications, yet existing benchmarks inadequately capture real-world clinical complexity. We introduce MEDSYN, a multilingual, multimodal benchmark of highly complex clinical cases with up to 7 distinct visual clinical evidence (CE) types per case. Mirroring clinical workflow, we evaluate 18 MLLMs on differential diagnosis (DDx) generation and final diagnosis (FDx) selection. While frontier models often match or even outperform human experts on DDx generation, all MLLMs exhibit a much larger DDx--FDx performance gap compared to expert clinicians, indicating a failure mode in synthesis of heterogeneous CE types. Ablations attribute this failure to (i) overreliance on less discriminative textual CE (\emph{e.g.}, medical history) and (ii) a cross-modal CE utilization gap. We introduce \emph{Evidence Sensitivity} to quantify the latter and show that a smaller gap correlates with higher diagnostic accuracy. Finally, we demonstrate how it can be used to guide interventions to improve model performance. 
\end{abstract}

\hspace{.5em}\includegraphics[width=1.25em,height=1.25em]{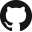}\hspace{.75em}%
\parbox{\dimexpr\linewidth-2\fboxsep-2\fboxrule}{%
  \href{https://github.com/jianing-lab/MEDSYN}{%
    \texttt{https://github.com/jianing}\linebreak\texttt{-lab/MEDSYN}%
  }%
}

\section{Introduction}
\begin{figure}[t]
    \centering

    \begin{subfigure}[t]{\columnwidth}
        \centering
        \includegraphics[width=0.9\linewidth]{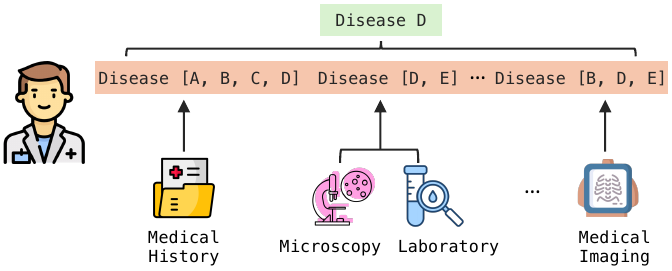}
        \caption{}
        \label{fig:teaser-a}
    \end{subfigure}


    \begin{subfigure}[t]{\columnwidth}
        \centering
        \includegraphics[width=0.9\linewidth]{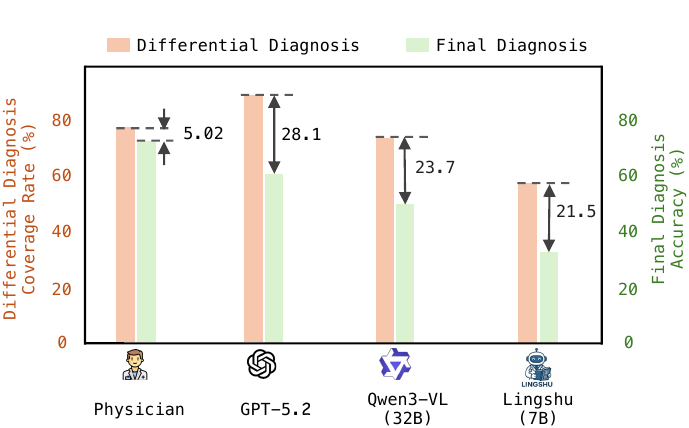}
        \caption{}
        \label{fig:teaser-b}
    \end{subfigure}

    \caption{\textbf{(a)} Clinicians curate a broad differential diagnosis (DDx) list before determining a final diagnosis (FDx) via evidence synthesis. \textbf{(b)} Models exhibit a substantial gap between DDx coverage rate and FDx accuracy, far exceeding that observed in human experts.}
    \label{fig:teaser}
\end{figure}

Multimodal Large Language Models (MLLMs) have demonstrated great potential in advancing clinical applications~\cite{qiu2024application}, yet the benchmarks used to evaluate them remain limited and fragmented. Early benchmarks~\cite{lau2018dataset, abacha2024vqa, liu2021slake, zhang2023pmc, hu2024omnimedvqa, ye2024gmai} primarily target single-image visual question answering (VQA) such as basic object recognition. More recent efforts~\cite{zuo2025medxpertqa, yu2025medframeqa} move toward more realistic settings by requiring integrative reasoning over multiple images, but several limitations persist. First, despite including multiple images, individual questions draw images from the same clinical-evidence (CE) type, \emph{e.g.}, cross-sectional CT scans. In clinical practice, clinicians typically leverage heterogeneous CE types, spanning laboratory tests, imaging from multiple modalities, microscopy images, and even omics data. This is particularly relevant for accurate diagnosis in complex clinical cases such as multimorbidity. While MedXpertQA MM~\cite{zuo2025medxpertqa} contains a multi-CE subset, it constitutes only a small fraction of the benchmark, with an average of 2.74 CE types per case. Second, most benchmarks emphasize selecting a final diagnosis (FDx). However, real diagnostic workflows typically begin with generating a differential diagnosis (DDx), \emph{i.e.}, a set of plausible conditions consistent with findings from one or more CE types, and then determine the FDx by synthesizing evidence across all available CE types. Finally, most existing benchmarks are English-only, limiting the assessment of MLLMs’ multilingual capabilities in clinical contexts.

To address these limitations, we introduce MEDSYN, a multilingual, multimodal benchmark of complex clinical cases, where each question contains, on average, \textbf{3.97 CE types} and \textbf{8.42 images} (Table~\ref{tab-1}) drawn from up to \textbf{7 distinct CE types} (Figure~\ref{fig:stats-evidence-per-case}). Mirroring real-world diagnostic workflows, our benchmark evaluates models on two tasks: (i) \textbf{DDx generation} and (ii) \textbf{FDx selection}. We benchmark 18 state-of-the-art MLLMs, including both general-purpose models (proprietary and open-source) and domain-specific medical models, with open-source scales ranging from 2B to 78B parameters. We further conduct two ablation studies using two medical MLLMs and their base model: (i) perturb textual CE by removing it or replacing it with length-matched random token strings; and (ii) run a leave-one-out analysis where each CE type is withheld and measure the resulting update in the model’s answer posterior, comparing the same CE provided as raw images versus expert-derived diagnostic findings from them. Our key findings are summarized as follows:


\begin{itemize}


    \item[\circled{1}] We show that leading models outperform expert clinicians on DDx generation but underperform on FDx selection, suggesting a capability gap between identifying plausible conditions from heterogeneous CE types and synthesizing them into a singular, accurate diagnosis. This gap varies by language, highlighting concerning cross-lingual disparities;

    \item[\circled{2}] We find that MLLMs overrely on textual inputs, skewing evidence weighting toward less discriminative CE (\emph{e.g.}, medical history). Removing such evidence increases attention to image tokens and, counterintuitively, improves diagnostic accuracy despite less CE being available;
    
    \item[\circled{3}] We demonstrate that visual understanding remains a major bottleneck: cross-modal misalignment distorts how MLLMs calibrate different CE types, yielding a \emph{cross-modal CE utilization gap}. We introduce a novel metric, termed \emph{Evidence Sensitivity}, to quantify this gap and show that a smaller gap correlates with higher diagnostic accuracy. We further show that this metric provides actionable guidance for targeted interventions to improve model performance.
\end{itemize}

\section{Related Work}

\begin{table}[t]
\centering
\tiny
\resizebox{\columnwidth}{!}{%
\begin{tabular}{lcccc}
\toprule
\textbf{Benchmark} &
\textbf{\# Images} &
\makecell[c]{\textbf{Multi-}\\\textbf{lingual}} &
\makecell[c]{\textbf{Avg. \# Images}\\\textbf{per Case}} &
\makecell[c]{\textbf{Avg. \# Evidence}\\\textbf{Types per Case}} \\
\midrule
VQA-Rad               & $204$        & \xmark & $1$ & $1$ \\
VQA-Med               & $500$        & \xmark & $1$ & $1$ \\
Slake                 & $96$         & \cmark & $1$ & $1$ \\
PMC-VQA               & $29$k        & \xmark & $1$ & $1$ \\
OmniMedVQA            & $118$k       & \xmark & $1$ & $1$ \\
GMAI-MMBench          & $21$k        & \xmark & $1$ & $1$ \\
MedXpertQA MM         & $2.8$k       & \xmark & $2.1$ & $2.74$ \\
\midrule
\textbf{MEDSYN}         & $3.6$k       & \cmark & $8.42$ & $3.97$  \\
\bottomrule
\end{tabular}
}
\caption{Comparison of MEDSYN with existing multimodal medical benchmarks.}
\label{tab-1}
\end{table}

\paragraph{Multimodal Large Language Models.} 
General-purpose MLLMs~\cite{openai2025introducinggpt5, anthropic2025claude4, comanici2025gemini} have demonstrated remarkable zero-shot capabilities on medical tasks, including diagnosing complex clinical cases~\cite{lam2025physician}, owing to the extensive clinical knowledge encoded in their LLM backbones~\cite{singhal2023large}. To further improve performance, recent work has explored domain-specific adaptation~\cite{chen2024huatuogpt, xu2025lingshu} by fine-tuning MLLMs on specialized medical data spanning diverse CE types. Despite these advances, current MLLMs remain prone to hallucinations and biases that impede safe deployment in real-world clinical environments~\cite{cross2024bias}. These challenges highlight the need for comprehensive benchmarks that evaluate models’ ability in complex and realistic diagnostic settings.

\paragraph{Multimodal Medical Benchmarks.} 
Existing medical benchmarks fall short of reflecting real-world clinical demands. Early benchmarks primarily target single-image VQA in narrow domains such as radiology~\cite{lau2018dataset, abacha2024vqa, liu2021slake} and pathology~\cite{he2020pathvqa}. Recently, more generalized benchmarks such as PMC-VQA~\cite{zhang2023pmc}, OmniMedVQA~\cite{hu2024omnimedvqa}, and GMAI-MMBench~\cite{ye2024gmai} have been proposed to assess MLLMs capabilities across diverse CE types. For instance, OmniMedVQA~\cite{hu2024omnimedvqa} covers 12 CE types, including multiple imaging modalities (\emph{e.g.}, CT, MRI, X-ray, ultrasound), microscopy, and specialized imaging such as colonoscopy. However, individual questions in these benchmarks remain isolated single-image snapshots tied to a single CE type. Although MedXpertQA MM~\cite{zuo2025medxpertqa} introduces multi-image, multi-CE VQA, this constitutes a small fraction of the benchmark and includes a limited number of CE types per question. Finally, most benchmarks are English-only, limiting evaluation of multilingual capabilities of MLLMs in clinical contexts. A detailed comparison is provided in Table~\ref{tab-1}.

\section{Benchmark}

\begin{figure*}[t]
    \centering
    \includegraphics[width=\textwidth]{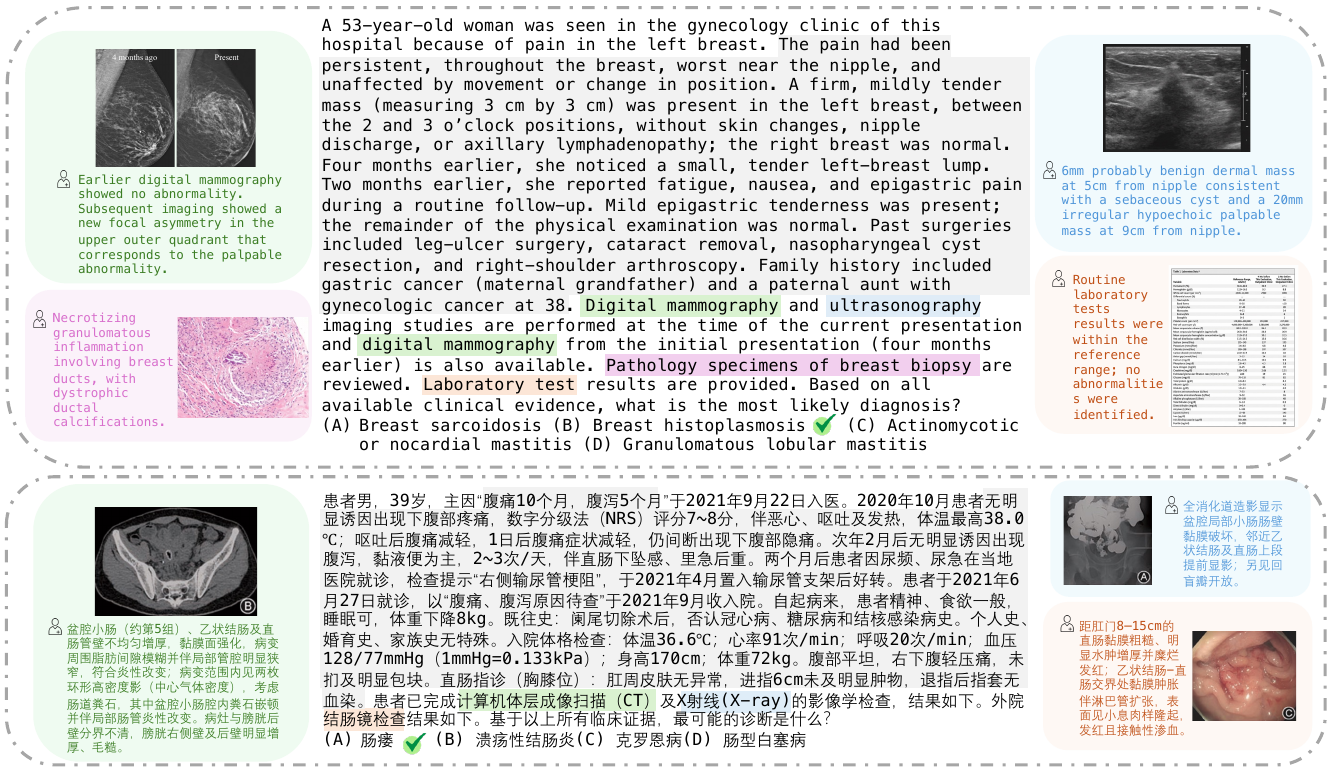}
    \caption{\textbf{Example final diagnosis selection tasks in English (top) and Chinese (bottom)}. Colors mark different visual clinical evidence (CE) types referenced in the question, with corresponding expert-derived diagnostic findings; gray denotes textual CE. In our experiment, each CE type is input as either raw images or text findings, not both.}
    \label{fig:example}
\end{figure*}

\subsection{Data Collection and Preprocessing}
\begin{figure}[t]
    \centering

    \begin{subfigure}[t]{\columnwidth}
        \centering
        \includegraphics[width=0.8\linewidth]{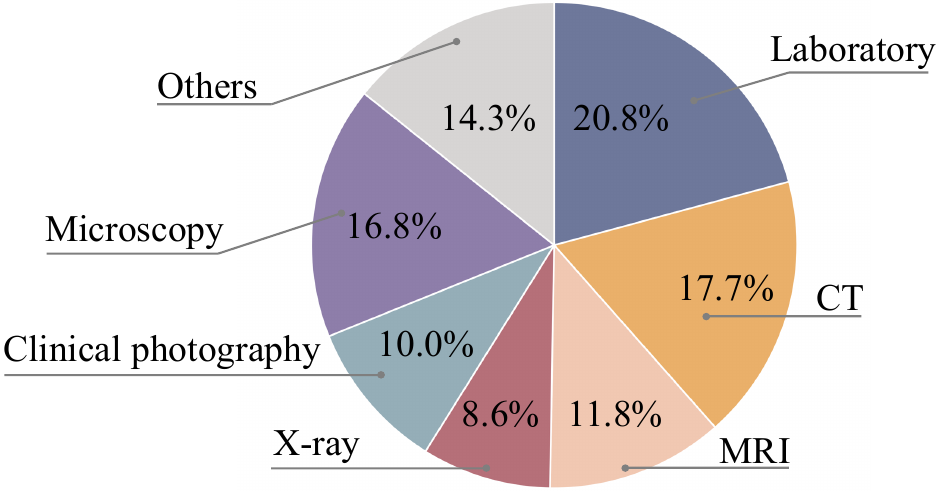}
        \caption{}
        \label{fig:stats-evidece-distribution}
    \end{subfigure}


    \begin{subfigure}[t]{\columnwidth}
        \centering
        \includegraphics[width=0.8\linewidth]{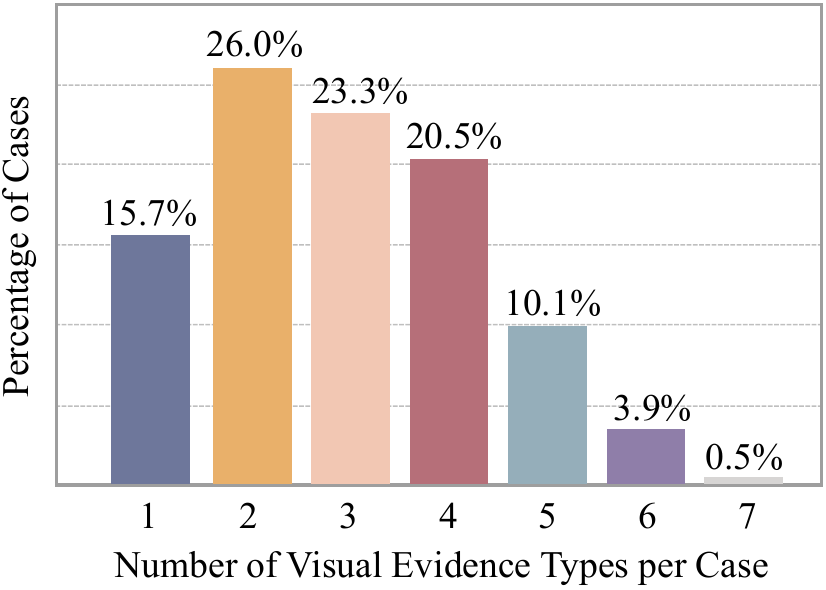}
        \caption{}
        \label{fig:stats-evidence-per-case}
    \end{subfigure}

    \caption{\textbf{(a)} Distribution of visual clinical evidence (CE) types. \textbf{(b)} Number of visual CE types per case.}
    \label{fig:stats}
\end{figure}

\paragraph{Data Collection.}
We collect consecutive English cases from the \emph{New England Journal of Medicine Case Record} series and Chinese case reports published in the \emph{National Medical Journal of China} (see Appendix~\ref{supp:case-example} for examples) from November 2015 to October 2025. For each case, we extract background and initial discussion up to DDx, along with all referenced tables and figures. In particular, the background contains textual CE including medical history and physical examination findings. The referenced tables and figures comprise visual CE from diagnostic investigations (\emph{i.e.}, objective, technical tests ordered by clinicians), such as laboratory studies, diagnostic imaging (\emph{e.g.}, CT, MRI and X-ray), microscopy and electrophysiological measurements (\emph{e.g.}, EEG). Finally, the initial discussion presents the expert-derived diagnostic findings from visual CE. The ground-truth (GT) FDx is taken from the FDx section or, if not explicitly stated, determined by a clinician based on the full report. All collected cases are subject to expert validation, and cases are excluded if (i) any individual visual CE is not diagnostically interpretable (\emph{e.g.}, low image quality, incomplete anatomical coverage, or artifacts), or (ii) the complete set of CE is diagnostically insufficient (\emph{i.e.}, the clinician is not able to confirm the FDx from all provided evidence). After validation, we obtain 452 cases in total (398 in English and 54 in Chinese). Figure~\ref{fig:stats} summarizes visual CE type distribution (Figure~\ref{fig:stats-evidece-distribution}) and counts per case (Figure~\ref{fig:stats-evidence-per-case}).

\paragraph{Data Preprocessing.}
For each case, we manually crop and label images by their CE type. For the same evidence from different time points, we add an additional reference when it is obtained. Image captions are added to the discussion, from which we employ GPT-5~\cite{openai2025introducinggpt5} to extract and organize expert-derived diagnostic findings based on CE types through in-context learning~\cite{dong2024survey}. This establishes a one-to-one mapping from visual CE to textual interpretation. Since a single CE can be reviewed by multiple clinicians, we utilize GPT-5 to summarize individual interpretations into a singular, cohesive summary to mitigate redundancy. The template prompts and examples of processed data are in Appendix~\ref{supp-data}. Finally, we conduct a manual quality check by cross-referencing the processed data with the original reports to ensure completeness and mitigate hallucinations, and a randomly selected 20\% subset is further validated by clinicians. 

\subsection{Benchmark Design}
To mirror real-world clinical workflow, we design two distinct tasks: (i) DDx generation, which involves information gathering and hypothesis generation, and (ii) FDx selection, which requires evidence synthesis and diagnostic verification. 

\paragraph{DDx generation.}
The DDx generation is an open-ended generation task where MLLMs are asked to present a differential list ranked from the most probable to least probable diagnosis. 

\paragraph{FDx selection.}
For FDx selection, we construct closed-ended multiple-choice questions (MCQs) where the single correct answer is the GT FDx and distractors are drawn from the model-generated DDx. We employ GPT-5 to generate a differential list and then select distractors according to two criteria: (i) distractors must exclude the FDx and any synonym or wording variant thereof; (ii) distractors should be clinically proximate to the FDx (\emph{e.g.}, a closely related histologic subtype; see Appendix~\ref{supp-data} for details). We apply an adversarial refinement process where we iteratively identify and eliminate potential shortcuts in how models distinguish correct from incorrect diagnoses~\cite{burgess2025microvqa}. Finally, the MCQs are reviewed by expert clinicians based on content validity and clinical relevance to ensure that discriminative CE explicitly disqualifies all distractors, thereby eliminating diagnostic ambiguity among the options. 

Figure~\ref{fig:example} illustrates two cases from FDx selection tasks with color-coded CE types. Each visual CE is paired with an expert-derived summary of diagnostic findings from it. During inference, a given CE type is provided as either visual or textual input, not both. Examples of DDx generation tasks are shown in Figure~\ref{fig-supp:example} in Appendix~\ref{supp:example}.

\paragraph{Metrics.} 
For DDx generation, we follow the evaluation framework of Kanjee et al.~\cite{kanjee2023accuracy} and employ GPT-5 as an automated judge to score generated DDx on a 0--5 scale (details in Appendix~\ref{supp-metric}). We report the coverage rate, defined as the percentage of cases whose DDx include the FDx, considering scores $\geq 4$ (\emph{i.e.}, DDx comprises the exact FDx or a highly synonymous condition) as a positive coverage based on clinician's suggestion. For FDx selection, we report overall accuracy.

\section{Experiments}
\subsection{Evaluation}
We evaluate a diverse set of MLLMs. Proprietary models include representative GPT~\cite{openai2025introducinggpt5}, Gemini~\cite{comanici2025gemini}, and Claude 4.5 Opus~\cite{anthropic2025claude4}. Open-source models span 2B–72B parameters, covering widely used families such as Qwen~\cite{yang2025qwen3}, InternVL~\cite{chen2024internvl}, and DeepSeek-VL~\cite{wu2024deepseek}. We also include three medical MLLMs: HuatuoGPT~\cite{chen2024huatuogpt}, Lingshu~\cite{xu2025lingshu}, and Med-Mantis~\cite{yang2025medical}. The evaluation is conducted using the VLMEvalKit~\cite{duan2024vlmevalkit} framework on 8 NVIDIA A6000 GPUs. We evaluate all models using a zero-shot setting. We also recruit two senior physicians to evaluate the English subset of the benchmark (see Appendix~\ref{supp:human} for details). 

\subsection{Main Results}

\begin{table*}[t]
\centering
\setlength{\tabcolsep}{6pt}
\definecolor{lightgrey}{gray}{0.93}
\resizebox{\textwidth}{!}{
\begin{tabular}{l l
| c >{\columncolor{lightgrey}}c
| c >{\columncolor{lightgrey}}c
| c >{\columncolor{lightgrey}}c}
\toprule
\multirow{2}{*}{\textbf{Type}} & \multirow{2}{*}{\textbf{Model}} &
\multicolumn{2}{c|}{\textbf{English}} &
\multicolumn{2}{c|}{\textbf{Chinese}} &
\multicolumn{2}{c}{\textbf{Overall}} \\
\cmidrule(lr){3-4}\cmidrule(lr){5-6}\cmidrule(lr){7-8}
& & \makecell{\textbf{DDx Coverage}\\\textbf{Rate (\%)}} 
& \makecell{\textbf{FDx Selection}\\\textbf{Acc. (\%)}} 
& \makecell{\textbf{DDx Coverage}\\\textbf{Rate (\%)}} 
& \makecell{\textbf{FDx Selection}\\\textbf{Acc. (\%)}} 
& \makecell{\textbf{DDx Coverage}\\\textbf{Rate (\%)}} 
& \makecell{\textbf{FDx Selection}\\\textbf{Acc. (\%)}} \\
\midrule

Expert & \textsc{Senior Physician}        & 77.13 & 72.11 &  -    &  -    &  -    &  -    \\
\midrule

\multirow{5}{*}{Proprietary} 
& \textsc{Claude Opus 4.5}         & 84.92 & \textbf{64.57} & 61.11 & \textbf{55.56} & 82.08 & \textbf{63.49} \\
& \textsc{GPT-O3}                  & \underline{86.15} & \underline{64.07} & \underline{64.81} & 48.15 & \underline{83.60} & \underline{62.17} \\
& \textsc{Gemini 2.5 pro}          & 80.60 & 62.81 & \underline{64.81} & \underline{53.70} & 78.72 & 61.72 \\
& \textsc{GPT-5.2}                 & \textbf{88.55} & 60.45 & \textbf{66.04} & \textbf{55.56} & \textbf{85.86} & 59.87 \\
& \textsc{Gemini 2.5 Flash}        & 73.55 & 57.54 & 54.72 & 37.04 & 70.50 & 55.09 \\
\midrule

\multirow{17}{*}{Open-Source}
& \multicolumn{7}{l}{\textit{Large}} \\
\cmidrule{2-8}
& \textsc{InternVL3 (78B)}         & 66.25 & 44.72 & 38.89 & 31.48 & 61.63 & 43.14 \\
& \textsc{Qwen2.5-VL (72B)}        & \underline{70.03} & \underline{48.24} & \textbf{51.85} & \textbf{40.74} & 63.94 & \underline{47.34} \\
\cmidrule{2-8}
& \multicolumn{7}{l}{\textit{Medium}} \\
\cmidrule{2-8}
& \textsc{InternVL3.5 (38B)}       & 64.23 & 43.47 & 37.74 & 24.07 & 53.53 & 41.15 \\
& \textsc{Qwen3-VL (32B)}          & \textbf{73.55} & \textbf{49.87} & \underline{48.00} & 33.33 & \textbf{70.50} & \textbf{47.89} \\
& \textsc{DeepSeek-VL2 (27B)}      & 55.67 & 32.91 & 25.93 & \underline{37.04} & 38.81 & 33.40 \\
\cmidrule{2-8}
& \multicolumn{7}{l}{\textit{Small}} \\
\cmidrule{2-8}
& \textsc{Qwen3-VL (8B)}           & 62.47 & 40.20 & 39.22 & 31.48 & \underline{66.08} & 39.16 \\
& \textsc{InternVL3.5 (8B)}        & 59.45 & 38.19 & 35.19 & 25.93 & 51.91 & 36.73 \\
& \textsc{Med-Mantis (8B)}         & 47.61 & 27.89 &  9.26 & 25.93 & 27.79 & 27.66 \\
& \textsc{HuatuoGPT-Vision (7B)}   & 60.71 & 38.69 & 24.07 & 31.48 & 49.01 & 37.83 \\
& \textsc{Lingshu (7B)}            & 58.44 & 36.93 & 22.22 & 33.33 & 49.90 & 36.50 \\
& \textsc{Qwen2.5-VL (7B)}         & 55.92 & 33.92 & 22.22 & 29.63 & 52.24 & 33.41 \\
\cmidrule{2-8}
& \multicolumn{7}{l}{\textit{Tiny}} \\
\cmidrule{2-8}
& \textsc{DeepSeek-VL2 (3B)}       & 51.89 & 30.90 & 20.37 & 24.07 & 21.46 & 30.08 \\
& \textsc{Qwen3-VL (2B)}           & 52.14 & 30.90 & 20.37 & 25.93 & 34.82 & 30.31 \\
\bottomrule
\end{tabular}
}
\caption{Comparison of differential diagnosis (DDx) coverage rate (\%) and final diagnosis (FDx) selection accuracy (\%) across models on the English and Chinese subsets, and overall. Best and second-best scores are in \textbf{bold} and \underline{underlined}, computed separately for proprietary and open-source models. Open-source models are grouped by parameter scale into \textit{Large} ($\approx$70B), \textit{Medium} ($\approx$30B), \textit{Small} ($\approx$7B), and \textit{Tiny} ($\approx$3B) categories. Clinical evidence from diagnostic investigations is provided as raw images.}
\label{tab-2}
\end{table*}

\begin{figure}[t]
    \centering
    \includegraphics[width=0.9\columnwidth]{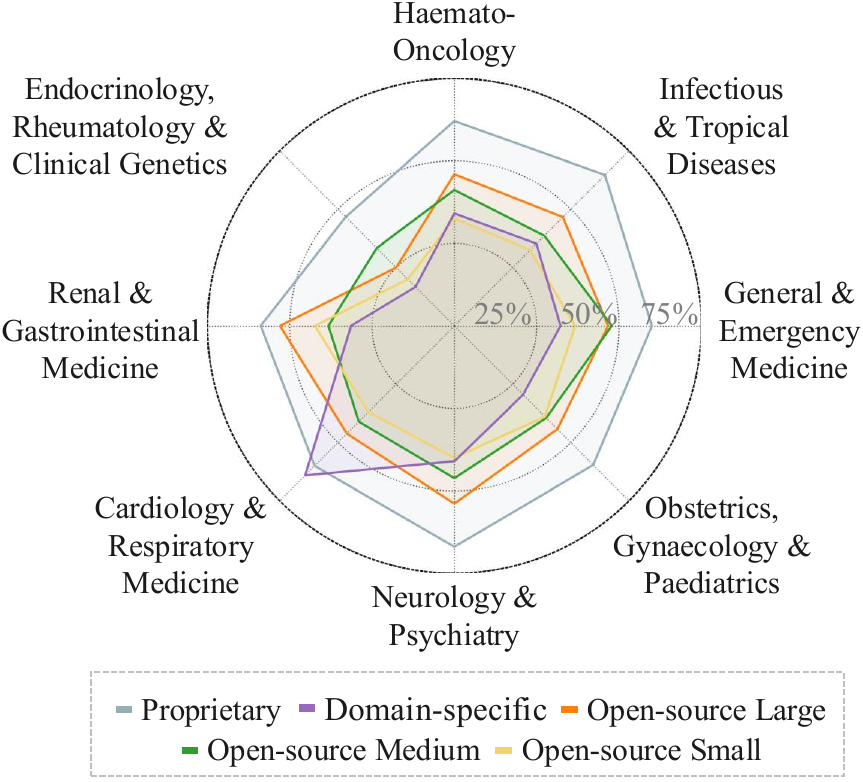}
    \caption{Mean final diagnosis accuracy (\%) for proprietary models, domain-specific models, and open-source general-purpose models across clinical specialties.}
    \label{fig:speciality}
\end{figure}

Table~\ref{tab-2} compares the performance of 18 MLLMs and senior physicians on DDx generation and FDx selection. Figure~\ref{fig:speciality} further details FDx performance across eight clinical specialties on the English subset. Overall, proprietary models consistently outperform open-source models, with an average gain of over 10 percentage points (pp) on both tasks. Claude Opus 4.5 achieves the highest FDx accuracy, while GPT-5.2 performs best on DDx generation. Among open-source models, Qwen3-VL (32B) demonstrates superior performance on both tasks, even surpassing models with significantly larger number of parameters. Among comparable settings, the domain-specific medical MLLMs, HuatuoGPT-Vision and Lingshu, both improve FDx accuracy over their base model, Qwen2.5-VL (7B). Notably, senior physicians achieve a mean accuracy of 72.11\% on FDx selection (English subset), which is 7.54 pp higher than the best MLLM, but score about 10 pp lower than the best model on DDx generation. Finally, individual model rankings for FDx selection vary between languages, indicating potential cross-lingual performance disparities.

Our key findings are as follows.

\paragraph{MLLMs Exhibit a Substantial Gap Between DDx Coverage and FDx Accuracy.} 
Across models, we observe a large gap between DDx coverage rate and FDx accuracy (approximately 20 pp on average on the English subset), far exceeding that of senior physicians (5.02 pp). The gap becomes even more pronounced for smaller models. For example, For example, HuatuoGPT-Vision achieves 60.71\% DDx coverage rate but only 38.69\% FDx accuracy. This suggests that while MLLMs can effectively enumerate potential diagnoses, they cannot yet function at the level of human experts in calibrating and synthesizing heterogeneous CE to correctly arrive at the FDx. On the Chinese subset, a similar trend can be observed for larger models ($\geq$70B), whereas small and medium-sized models show a divergent trend, sometimes exhibiting a negative gap. This shift, however, likely reflects a floor effect: since lower-tier models perform poorly on Chinese tasks, their results are often close to random guessing, so the gaps are largely by chance.

\paragraph{Visual Understanding Remains a Diagnostic Bottleneck.} 
Table~\ref{tab-3} reports the accuracy of FDx when raw visual evidence from diagnostic investigations is replaced by expert interpretations. As shown in Table~\ref{tab-3}, both proprietary and open-source models demonstrate a significant accuracy gain of over 10 pp, suggesting that the vision encoder’s inability to accurately capture granular details in complex clinical imagery remains a primary bottleneck for MLLMs in diagnostic tasks.

\begin{table}[htbp]
\centering
\setlength{\tabcolsep}{7pt}
\renewcommand{\arraystretch}{1.15}
\resizebox{\columnwidth}{!}{\begin{tabular}{lcc}
\toprule
\textbf{Model} & \textbf{Raw Images} & \makecell{\textbf{Expert-Derived}\\\textbf{Diagnostic Findings}} \\
\midrule
\textsc{GPT-O3}                           & 64.07 & 75.38 {\scriptsize(+11.31)} \\
\textsc{GPT-5.2}                          & 60.45 & 75.13 {\scriptsize(+14.68)} \\
\midrule
\textsc{Qwen2.5-VL (72B)}                 & 48.24 & 65.83 {\scriptsize(+17.59)} \\
\textsc{Qwen3-VL (32B)}                   & 49.87 & 65.58 {\scriptsize(+15.71)} \\
\textsc{InternVL3 (78B)}                  & 44.72 & 63.82 {\scriptsize(+19.10)} \\
\textsc{Qwen3-VL (8B)}                    & 40.20 & 59.05 {\scriptsize(+18.85)} \\
\textsc{Lingshu (7B)}                     & 36.93 & 54.27 {\scriptsize(+17.34)} \\
\textsc{HuatuoGPT-Vision (7B)}            & 38.69 & 52.26 {\scriptsize(+13.57)} \\
\textsc{Qwen2.5-VL (7B)}                  & 33.92 & 50.50 {\scriptsize(+16.58)} \\
\bottomrule
\end{tabular}}
\caption{Final diagnosis accuracy (\%) across models on English subset. We compare performance when clinical evidence from diagnostic investigations is provided as raw images versus expert-derived text findings.}
\label{tab-3}
\end{table}


\paragraph{Domain-Specific Training Can Outperform Parameter Scale in Specialized Clinical Tasks.}
We analyze FDx performance across 8 clinical specialties on the English subset. We report mean accuracy for proprietary models, domain-specific models, and open-source general-purpose models grouped by size: large ($\approx$70B), medium ($\approx$30B), and small ($\approx$7B). As shown in Figure~\ref{fig:speciality}, proprietary models exhibit the most balanced performance across the clinical spectrum, notwithstanding a systematic performance deficit in obstetrics, gynaecology, and paediatrics. Notably, medical domain-specific models achieved a mean accuracy of 64.10\% in cardiology and respiratory medicine, outperforming both proprietary models (60.00\%) and general-purpose models with tenfold larger parameter counts (46.15\%). We hypothesize that this localized dominance stems from specialized physiological mappings inherent to these disciplines such as correlation between ECG morphology and cardiac conduction events, which domain-specific finetuning more effectively captures than simple parameter scaling.

\subsection{Analytical Results}
\label{ana_results}
\finding{Finding 1: Text bias miscalibrates evidence weighting in MLLMs.}

Recent works show that MLLMs frequently exhibit a strong bias toward linguistic signals, placing “blind faith” in text even when visual evidence is presented~\cite{deng2025words, lee2025vlind}. In clinical practice, textual CE such as medical history and physical findings is essential for DDx generation but often contains nonspecific features that are common to multiple diagnoses. Conversely, visual CE from diagnostic investigations (\emph{e.g.}, diagnostic imaging and microscopy) is typically more discriminative, increasing physician confidence in FDx by over 30\%~\cite{peterson1992contributions}.

\begin{figure}[htbp]
    \centering
    \includegraphics[width=\linewidth]{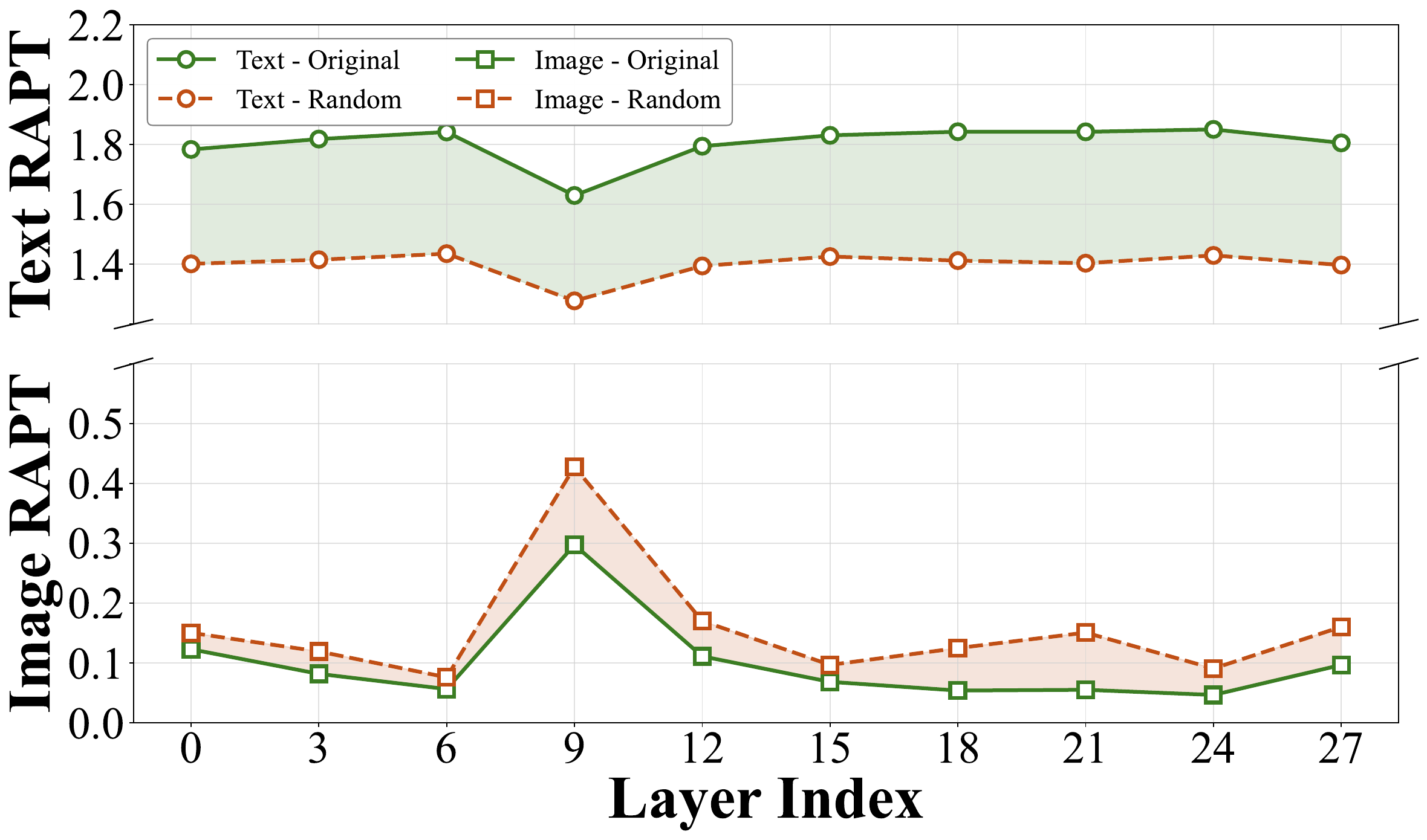}
    \caption{Layer-wise Relative Attention per Token (RAPT) for Lingshu on text (excluding question stem) versus image tokens, before and after the \textsc{Random-Text} intervention.}
    \label{fig:rapt-lingshu-maintext}
\end{figure}

To investigate how this bias affects MLLM evidence calibration, we conducted two ablation studies on the English subset of the FDx selection task: (i) \textsc{Remove-Text}, where textual CE is omitted, and (ii) \textsc{Random-Text}, where textual CE is replaced by length-matched, randomly sampled tokens to isolate the effect of token budget from semantic content. As shown in Table~\ref{tab-4}, both ablations yield consistent performance gains across all three models, with larger improvements for domain-specific models. This counterintuitive \emph{less-is-more} effect that reducing available CE yet improves accuracy suggests textual CE acts as a distractor instead of helpful context for FDx selection due to inherent bias in MLLMs. For the \textsc{Random-Text} ablation, we further analyze the Relative Attention per Token (RAPT)~\cite{liu2025seeing} (see Appendix~\ref{supp:rapt} for formal definition) for Lingshu on text and image tokens across layers (see Appendix~\ref{supp:additional-results} for additional models). Our analysis (Figure~\ref{fig:rapt-lingshu-maintext}) reveals that while the model generally allocates disproportionately high attention to text than image tokens, this gap narrows under the \textsc{Random-Text} condition. In particular, we observe a more pronounced RAPT surge on image tokens within deeper layers which have been identified as visual grounders in recent work~\cite{liu2025seeing}. 

Overall, our results suggest MLLMs suffer from an evidence miscalibration failure mode where the models underweight visual CE despite its superior diagnostic value due to inherent bias toward text. Such bias is semantics-sensitive: coherent clinical text diverts attention away from visual grounding, whereas removing or randomizing it attenuates this effect and restores the model's focus to more informative visual features.

\begin{table}[t]
\centering
\tiny
\setlength{\tabcolsep}{4.5pt}
\renewcommand{\arraystretch}{1.15}
\resizebox{\columnwidth}{!}{
\begin{tabular}{lccc}
\toprule
\textbf{Setting} & \textbf{HuatuoGPT-Vision} & \textbf{Lingshu} & \textbf{Qwen2.5-VL (7B)} \\
\midrule
\textsc{Baseline}     & 38.69  & 36.93 & 33.92 \\
\midrule
\textsc{Remove-Text}  &
\cellcolor{green!17.5}42.21 {(+3.52\,pp)} &
\cellcolor{green!35.0}43.97 {(+7.04\,pp)} &
\cellcolor{green!2.5}34.42 {(+0.50\,pp)} \\
\textsc{Random-Text}  &
\cellcolor{green!5.0}39.70 {(+1.01\,pp)} &
\cellcolor{green!23.8}41.71 {(+4.78\,pp)} &
\cellcolor{green!3.7}34.67 {(+0.75\,pp)} \\
\bottomrule
\end{tabular}}
\caption{Final diagnosis accuracy (\%) before and after \textsc{Remove-Text} and \textsc{Random-Text} interventions.}
\label{tab-4}
\end{table}

\finding{
Finding 2: Cross-modal misalignment distorts how MLLMs calibrate evidence.
}
Recent studies reveal a misalignment between text and image modalities in MLLMs, where visual encoding introduce noise or information loss when mapping raw images to image tokens~\cite{jain2025words, shu2025large, li2025lost}. We hypothesize this gap can lead to inconsistent weighting of identical CE presented in different modalities. 

\begin{figure}[t]
    \centering

    \begin{subfigure}[t]{\columnwidth}
        \centering
        \includegraphics[width=\linewidth]{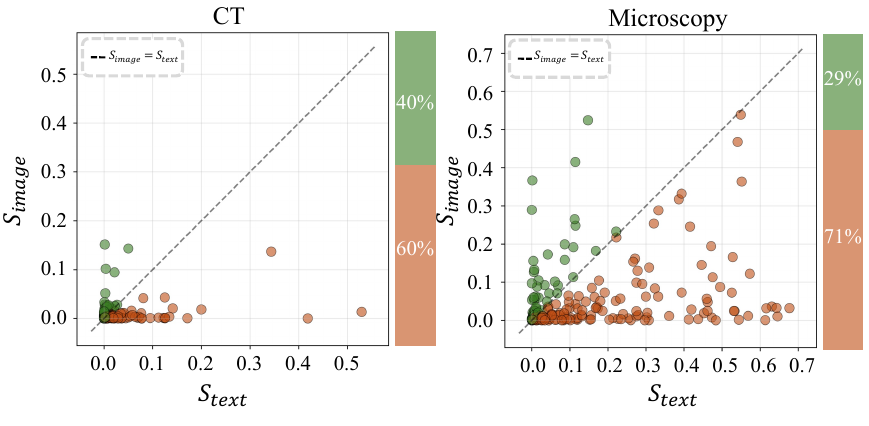}
        \caption{Qwen2.5-VL (7B)}
        \label{fig:rapt-qwen}
    \end{subfigure}


    \begin{subfigure}[t]{\columnwidth}
        \centering
        \includegraphics[width=\linewidth]{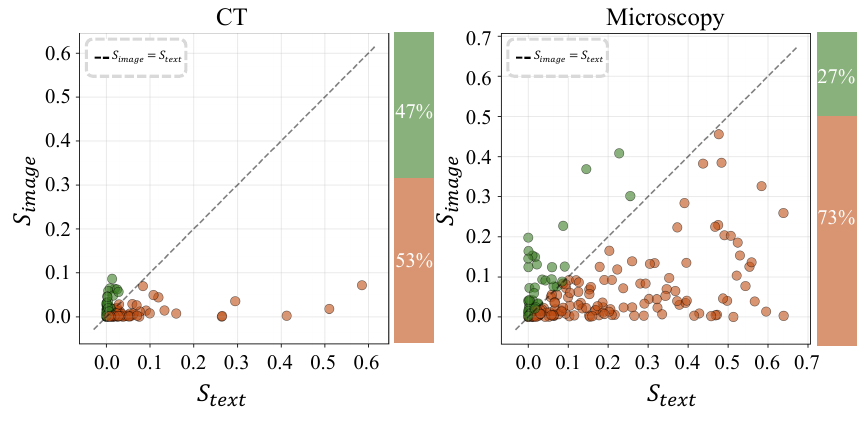}
        \caption{Lingshu (7B)}
        \label{fig:rapt-lingshu}
    \end{subfigure}

    \caption{Cross-modal sensitivity on CT and microscopy. \textcolor{forestgreen}{Green} and \textcolor{burntorange}{brown} points indicate cases where \textcolor{forestgreen}{$S^{(m)}{\text{image}}>S^{(m)}{\text{text}}$} and \textcolor{burntorange}{$S^{(m)}{\text{image}}<S^{(m)}{\text{text}}$}, respectively. The right-hand bar chart shows the proportion of cases in each category.}
    \label{fig:sensitivity}
\end{figure}

To investigate this hypothesis, we introduce \emph{Evidence Sensitivity} to measure the impact of specific CE on the model's decision-making. Formally, let $\mathcal{E}=\{e_1, \dots, e_M\}$ denote the set of CE for a given case, and let $p_\theta(y \mid \mathcal{E})$ represent the model's answer posterior over the answer space $y$. For each CE type $e_m\in\mathcal{E}$, we define its sensitivity as the shift in the model’s answer posterior upon its removal, computed as the Jensen--Shannon divergence (JSD) between the full and ablated distributions: 
\begin{equation}
S^{(m)} \;=\; \mathrm{JSD}\!\left(p_\theta(y \mid \mathcal{E}) \parallel\, p_\theta(y \mid \mathcal{E}\setminus \{e_m\})\right).
\end{equation}
This metric captures the update in model's belief when evidence $e_m$ is omitted. We compute the sensitivity for the same evidence presented in two distinct modalities: (i) raw image input ($S^{(m)}_{\text{image}}$) and (ii) expert-derived textual summaries of diagnostic findings from the image. Theoretically, identical evidence should yield comparable belief updates across modalities, \emph{i.e.}, $S^{(m)}_{\text{image}} \approx S^{(m)}_{\text{text}}$, with data points $(S^{(m)}_{\text{text}}, S^{(m)}_{\text{image}})$ clustering along the diagonal. Deviations from the diagonal define a \emph{cross-modal CE utilization gap}, reflecting the extent of evidence calibration distortion stemming from cross-modal misalignment. Figure~\ref{fig:sensitivity} provides a qualitative comparison between two CE types (\emph{i.e.,} CT and Microscopy) for Qwen2.5-VL (7B) and Lingshu (see Figure~\ref{fig-supp:sensitivity} and~\ref{fig-supp:percentage} in Appendix~\ref{supp:additional-results} for results on more CE types and for HuatuoGPT-Vision). From the plot we can observe data points rarely align with the diagonal. Instead, they predominantly cluster along either the horizontal or vertical axes, forming an L-shaped distribution. This pattern indicate a cross-modal distortion in evidence calibration: models remain relatively insensitive to evidence in one modality that they otherwise deem critical in another. Notably, for microscopy, most data points cluster below the diagonal (\emph{i.e.}, \textcolor{burntorange}{$S^{(m)}{\text{image}}<S^{(m)}{\text{text}}$}), suggesting a systematic undersensitivity to visual CE. 

We further quantify this distortion by computing the Normalized Mean Squared Error (NMSE) relative to the diagonal:
\begin{equation}
\mathrm{NMSE}_{y=x} \;=\;
\frac{\sum_{i=1}^{N} \left( S^{(m)}_{\text{image}, i} - S^{(m)}_{\text{text}, i} \right)^2}{\sum_{i=1}^{N} \left( S^{(m)}_{\text{text}, i} \right)^2}
\label{eq:nmse}
\end{equation}
where $i$ denotes individual cases and $N$ is the total number of cases. Table~\ref{tab-5} reports the NMSE for different evidence types across the three models (detailed results for each specific CE type are provided in Table~\ref{tab-supp-5} of Appendix~\ref{supp:additional-results}). Our analysis reveals that a lower mean $\mathrm{NMSE}_{y=x}$ correlates with higher overall accuracy.

\begin{table}[htbp]
\centering
\setlength{\tabcolsep}{4.5pt}
\renewcommand{\arraystretch}{1.15}
\resizebox{\columnwidth}{!}{\begin{tabular}{lcc}
\toprule
\textbf{Model} & \textbf{Average NMSE $\downarrow$} & \textbf{Acc. $\uparrow$} \\
\midrule
\textsc{Qwen2.5-VL (7B)}         & 0.98 & 33.92\\
\textsc{Lingshu (7B)}            & 0.95 & 36.93\\
\textsc{HuatuoGPT-Vision (7B)}   & \textbf{0.93} & \textbf{38.69}\\
\bottomrule
\end{tabular}} 
\caption{Comparison of average normalized mean squared error (NMSE) and final diagnosis accuracy (\%) across models. Best results are highlighted in \textbf{bold}.}
\label{tab-5}
\end{table}

Finally, we demonstrate that the cross-modal CE utilization gap is actionable through two strategies: (i) test-time prompt refinement and (ii) targeted supervised finetuning (SFT). 

\paragraph{Test-time Prompt Refinement.}
To address the model's systematic undersensitivity to microscopy images, we redesign the system prompt to explicitly instruct the model to attend to it (see Appendix~\ref{supp:prompt_refinement} for details). Following this intervention, we observe that, across models, the fraction of instances below the diagonal decreases (see Figure~\ref{fig-supp:prompt_refinement} in Appendix~\ref{supp:additional-results}), indicating increased sensitivity to microscopy images. Consistently, the cross-modal gap narrows, reflected by lower microscopy NMSE (see Table~\ref{tab-supp-6} in Appendix~\ref{supp:additional-results}). Crucially, the reduced gap translates directly to downstream performance. As shown in Table~\ref{tab-6}, overall accuracy improves across three models after prompt refinement. These findings underscore that bridging the utilization gap between visual and textual evidence is essential for achieving reliable, high-performance clinical decision-making.


\begin{table}[t]
\centering
\renewcommand{\arraystretch}{1.15}
\setlength{\tabcolsep}{4.5pt}
\resizebox{\columnwidth}{!}{
\begin{tabular}{lcc}
\toprule
\textbf{Model} & \textbf{Baseline Acc.} & \textbf{Prompt Refinement Acc.} \\
\midrule
\textsc{Qwen2.5-VL (7B)} &
33.92 &
\cellcolor{green!3.7}34.67 {(+0.75\,pp)} \\
\textsc{Lingshu} &
36.93 &
\cellcolor{green!7.4}38.44 {(+1.51\,pp)} \\
\textsc{HuatuoGPT-Vision} &
38.69 &
\cellcolor{green!1.2}38.94 {(+0.25\,pp)} \\
\bottomrule
\end{tabular}}
\caption{Comparison of final diagnosis accuracy (\%) before and after prompt refinement across models.}
\label{tab-6}
\end{table}

\paragraph{Targeted SFT.}
Another strategy to mitigate the utilization gap is to finetune the model on a curated dataset enriched with images from undersensitive CE types (\emph{i.e.}, targeted SFT). Specifically, we curate a \emph{microscopy-heavy} dataset comprising 50\% microscopy images and 50\% other medical images. As shown in Table~\ref{tab-7}, Qwen2.5-VL (7B) model finetuned on this curated dataset achieves an accuracy improvement of 1.51 pp, doubling the 0.75 pp gain from general SFT on non-curated data. These results demonstrate that cross-modal evidence utilization gap provides a strategic signal for guiding SFT data curation: by up-sampling CE types that are systematically underutilized by the base model, we can achieve better overall performance after SFT. We provide experiment details in Appendix~\ref{appendix:sft}.

\begin{table}[t]
\centering
\small
\setlength{\tabcolsep}{4.5pt}
\renewcommand{\arraystretch}{1.15}
\begin{tabular}{lc}
\toprule
\textbf{Setting} & \textbf{Acc.} \\
\midrule
\textsc{Baseline} & 33.92 \\
\midrule
\textsc{General SFT} &
\cellcolor{green!3.7}34.67 {(+0.75)} \\
\textsc{Targeted SFT} &
\cellcolor{green!7.4}\textbf{35.43} {(\textbf{+1.51})} \\
\bottomrule
\end{tabular}
\caption{Final diagnosis accuracy (\%) for Qwen2.5VL before and after supervised fine-tuning (SFT), comparing general with targeted SFT. Best results are in \textbf{bold}.}
\label{tab-7}
\end{table}

\section{Conclusion}
We introduce MEDSYN, a multilingual, multimodal benchmark of highly complex clinical cases that necessitate synthesis of multiple CE types for accurate diagnosis. We evaluate 18 MLLMs on both DDx generation and FDx selection tasks. Across models, we observe a critical gap between DDx coverage rate and FDx accuracy, substantially larger than that observed for human experts, indicating that MLLMs struggle to synthesize heterogeneous CE to arrive at the correct FDx. We further attribute it to two primary factors: overreliance on textual CE and cross-modal CE utilization gap. Together, these factors lead to an evidence-calibration failure mode, where models weight clinical evidence not only by its diagnostic value but also by their internal bias. We introduce Evidence Sensitivity to quantify this cross-modal gap, and show that a narrower gap correlates with better overall performance. Finally, we demonstrate that this gap provides actionable guidance for targeted interventions to improve model performance.

\section*{Limitations}
\paragraph{Linguistic Imbalance.} 
The Chinese subset constitutes only a smaller portion ($\sim$12\%) of the total benchmark. This disparity may limit the generalizability of our findings regarding model performance in non-English clinical contexts and diverse medical linguistic environments.

\paragraph{Absence of Chinese Physician Baseline.} 
While we provide human expert benchmarks for the English subset, we currently lack a comparable physician baseline for the Chinese subset. Without this comparison, it is difficult to fully contextualize the gap between DDx coverage and FDx accuracy for models specifically on Chinese clinical cases.

\paragraph{Automated Evaluator Bias.} 
Our evaluation framework utilizes GPT-5 as a judge to score generated DDx. While this automated approach enables scalable evaluation, it may introduce evaluator bias, potentially favoring models with similar linguistic patterns (\emph{e.g.}, GPT-5.2 and GPT-o3) or over-indexing on specific medical terminologies.

\paragraph{Influence of Image Type and Order.}
Our current evaluation inputs all images simultaneously. Consequently, it does not investigate how the specific types of imaging modalities included in a case impact the diagnostic performance of MLLMs, nor does it analyze how the input order of these images might affect a model's clinical reasoning.

\section*{Acknowledgement}
This research was primarily supported by the ETH AI Center through an ETH AI Center doctoral fellowship to Boqi Chen.

\bibliography{custom}

\appendix

\section{Details of Data Collection and Processing}
\begin{figure*}[t]
    \centering

    \begin{subfigure}[t]{0.55\textwidth}
        \centering
        \includegraphics[width=\linewidth]{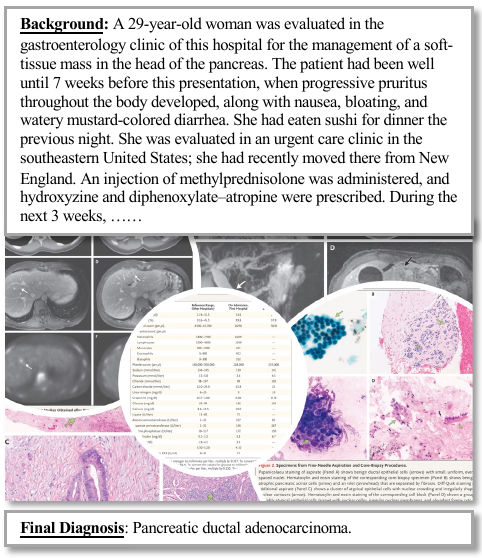}
        \caption{}
        \label{fig-case-report-eng}
    \end{subfigure}\hfill
    \begin{subfigure}[t]{0.44\textwidth}
        \centering
        \includegraphics[width=\linewidth]{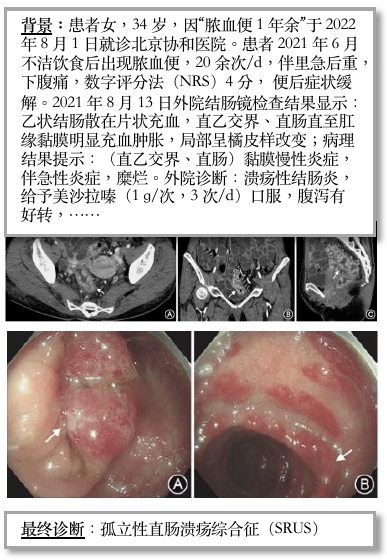}
        \caption{}
        \label{fig-case-report-chn}
    \end{subfigure}

    \caption{Examples of raw case reports. \textbf{(a)} A case record from Massachusetts General Hospital (Boston, MA); \textbf{(b)} A case report published by National Medical Journal of China .}
    \label{fig-supp:report}
\end{figure*}
\subsection{Examples of Case Reports}
\label{supp:case-example}
Figure~\ref{fig-supp:report} showcases two examples of case reports from the \emph{New England Journal of Medicine Case Record} series (Figure~\ref{fig-case-report-eng}) and the \emph{National Medical Journal of China} (Figure~\ref{fig-case-report-chn}). 

\subsection{Prompt for Data Processing}
\label{supp-data}
The prompt for extracting textual evidence and clinician’s interpretation of visual evidence based on evidence types:
\begin{lstlisting}[style=appendixprompt]
You are an clinical information extraction assistant.

Task
Given a single medical case consisting of:
(1) Case ID
(2) Background text
(3) Figure caption text (captions extracted from figures)
return exactly ONE flat JSON object with verbatim (copy-pasted) content.
Do not paraphrase.

Inputs
- Case ID: {case_id}
- Background: {background}
- Caption: {caption}

Output format
Return only valid JSON (no prose, no markdown code blocks).

Required keys
- "Case ID": string
- "Modality": string (comma-separated)

Optional keys (include only if present)
- "Patient Background"
- "Laboratory Findings"
- "Medications"
- "Impression / Differential" (ONLY if explicitly labeled)
- "CT Findings"
- "MRI Findings"
- "PET Findings"
- "Ultrasound Findings"
- "X-ray Findings"
- "Pathology Findings"
- "Clinical Photography Findings"
- "Other Imaging Findings" (include the modality name in the value)

Value rules (for ALL optional keys)
- Each value is a single string made of one or more verbatim chunks.
- Join chunks with a DOUBLE line break: \n\n
- Preserve internal line breaks within a chunk.
- Keep temporal/procedural qualifiers inside the chunk (e.g., "after
  contrast", "hospital day 2", "1 year earlier", planes/reconstructions).
- Deduplicate materially identical statements (keep the clearest).
- If duplicates conflict, keep both (each with its own source tag).

Source tags (append when available; do NOT invent)
Allowed tag formats:
- (Dr. <Name>)
- (Figure <N> caption) OR (Figure caption) if unnumbered
If no source is available, omit the tag.

Modality extraction (concise rules)
- Detect modalities from Caption first. If Caption is entirely absent,
  infer modalities from Background.
- Allowed modalities (for "Modality" and imaging keys):
  CT, MRI, PET, Ultrasound, X-ray, Pathology, Clinical Photography, Other Imaging.
- Hybrids (e.g., PET-CT): split into PET and CT when feasible; duplicate
  only if a sentence truly spans both.
- Assign each imaging sentence to its matching modality key. If ambiguous,
  use "Other Imaging Findings" and name the modality exactly as written.

Rule for the "Modality" key
- List ONLY modalities present in FIGURES (from Caption).
- If Caption is entirely absent, list modalities discussed in Background.
- Do NOT include non-imaging categories (e.g., labs, meds).

Caption handling
- If multiple figures are described, split by figure.
- Attribute each extracted chunk with the precise figure source tag.

Background handling
- Extract verbatim content for patient background, labs, medications, and any
  explicitly labeled impression/differential.
- Add imaging from Background ONLY if Caption is absent or clearly
  incomplete.
- Do not add interpretation unless "Impression / Differential" is
  explicitly labeled.

Hard constraints (do NOT do these)
- Do NOT paraphrase or normalize numbers/dates/units.
- Do NOT output nested JSON; output a single flat object.
- Do NOT include internal/system markers (e.g., :contentReference[...],
  oaicite).
- Do NOT use file names as sources.
- Do NOT output anything except the single JSON object.

Procedure
1. Parse Caption -> split per figure, identify modality cues, extract
   verbatim chunks with figure source tags.
2. Parse Background -> extract non-imaging (background, labs, meds) and any
   labeled impression/differential; add imaging only if needed.
3. Build "Modality" using the rule above.
4. Aggregate chunks per key with \n\n and return JSON only.

Here are some examples:
...
\end{lstlisting}

The prompt for summarizing individual expert interpretations:
\begin{lstlisting}[style=appendixprompt]
You are an expert medical summarization assistant.

Task
Given ONE case, produce concise summaries while preserving the case ID,
modality list, medications, and diagnosis information if present.
Use ONLY the provided text. Do not invent or infer missing details.

Inputs
- Case ID: {case_id}
- Modality: {modality_string}   (comma-separated; provided by dataset)
- Patient Background: {patient_background}
- Findings sections (zero or more):
  - {findings_key_1}: {findings_text_1}
  - {findings_key_2}: {findings_text_2}
  - ...

Outputs
Return only valid JSON of summaries with the exact formats below.

A) Patient background summary (if Patient Background is provided)
Instructions
- Use professional medical language.
- Remove any explicit diagnosis names already stated in the background.
- Remove any imaging or laboratory results described in the background
  (including impressions/findings/conclusions).
- Focus on: presenting problem and context, key past medical/psychiatric
  history, major exam findings, complications/risk factors, current status
  and plan (including disposition and follow-up).
- Do not include author names or citations in parentheses.

Output format (exact labels)
Patient Background Summary: <single concise paragraph>

B) Imaging (or other) findings summary (for EACH provided findings section)
For each input section "{modality_or_section_name} Findings":
Instructions
- Include clinically meaningful abnormalities and anatomic locations
  (use specific structures when stated).
- Include interpretation ONLY if explicitly stated in the source text.
- Include comparisons to prior studies when stated.
- Keep modality/technique details only if they affect meaning.
- Exclude demographics, physician names, and extraneous technical details.
- Output ONLY the summary paragraph for that section.

Output format (per section)
<{modality_or_section_name} Findings Summary>: <single concise paragraph>

Hard constraints
- Do NOT add new diagnoses or conclusions.
- Do NOT normalize or change numbers, dates, or units.
- Do NOT output anything beyond the specified formats.

Here are some examples:
...
\end{lstlisting}

The prompt for generating distractors for multiple-choice questions (MCQs):
\begin{lstlisting}[style=appendixprompt]
You are an expert clinician and adversarial exam-writer.

Goal
Given one case (history + imaging + labs) plus a final diagnosis and
a comma-separated differential list, produce three confusing but incorrect distractor diagnoses.

Inputs
- Patient history: {history_text}
- Imaging findings: {imaging_text}
- Laboratory findings: {labs_text}
- Ground-truth final diagnosis: {ground_truth_text}
- Differential diagnoses (comma-separated): {differential_diagnoses_text}

Core rules
1) Distractor constraints (must all be true)
- Output EXACTLY three distractors.
- ALL distractors MUST be drawn from the differential list.
  (You may normalize wording but must not change the underlying diagnosis.)
- Each distractor must be a real, recognized diagnosis.
- Each distractor must be genuinely plausible for THIS patient given the
  provided history, imaging, and labs.
- Do NOT output the ground truth or any synonym/trivial variant of it as a
  distractor.
- Avoid non-diagnoses (pure symptoms, risk factors, procedures).

2) Make distractors confusing
- Prefer distractors in the same organ system or a closely related one.
- Prefer distractors that share major features with the ground truth
  (symptoms, labs, imaging, histology).
- Each distractor should differ from the ground truth by subtle
  details (e.g., anatomic site, vascular territory, mechanism, histologic
  variant), such that there is still exactly ONE best answer.

3) Remove lexical giveaways
- Keep ground truth and distractors similar in "shape":
  same head noun/pattern when possible (e.g., all "... adenocarcinoma",
  all "... ischemic stroke", all "... cardiomyopathy").
- Keep length and syntactic structure similar across the four options.
- Do not let only one option contain a uniquely specific keyword unless
  the others contain parallel, equally specific keywords.

Output format (JSON only)
Return ONLY valid JSON. No explanations. No comments. No code fences.
Use this exact schema:
{
  "ground_truth": "<string: normalized primary diagnosis>",
  "distractors": ["<string>", "<string>", "<string>"]
}

Here are some examples:
...
\end{lstlisting}

\subsection{Examples of Differential Diagnosis Generation Cases}
\label{supp:example}
\begin{figure*}[htbp]
    \centering
    \includegraphics[width=0.95\textwidth]{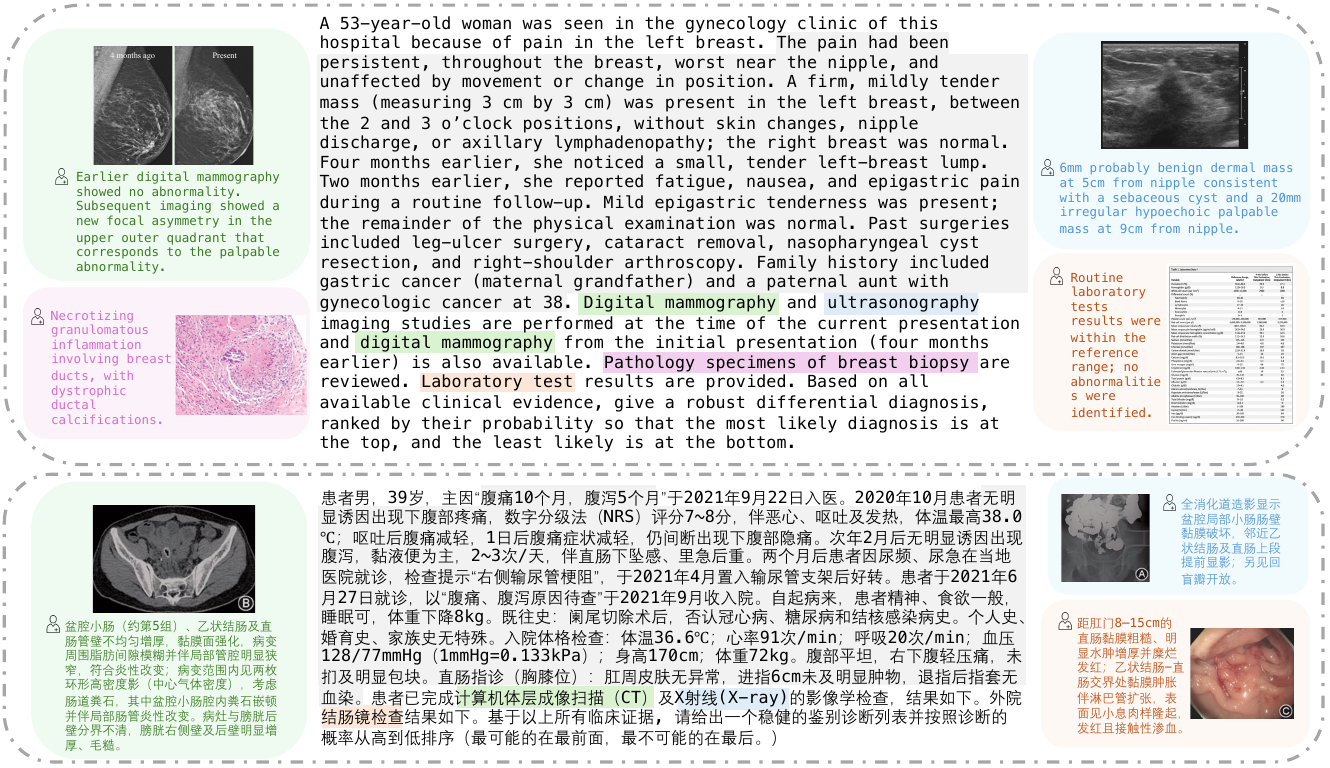}
    \caption{\textbf{Example differential diagnosis generation cases in English (top) and Chinese (bottom)}. Colors mark different visual clinical evidence (CE) types referenced in the question, with corresponding expert-derived diagnostic findings; gray denotes textual CE. In our experiment, each CE type is input as either raw images or text findings, not both.}
    \label{fig-supp:example}
\end{figure*}

Figure~\ref{fig-supp:example} provides two example cases of differential diagnosis (DDx) generation.

\subsection{Detailed Distribution of Clinical Specialties}
\begin{figure*}[htbp]
    \centering

    \begin{subfigure}[t]{\textwidth}
        \centering
        \includegraphics[width=\textwidth]{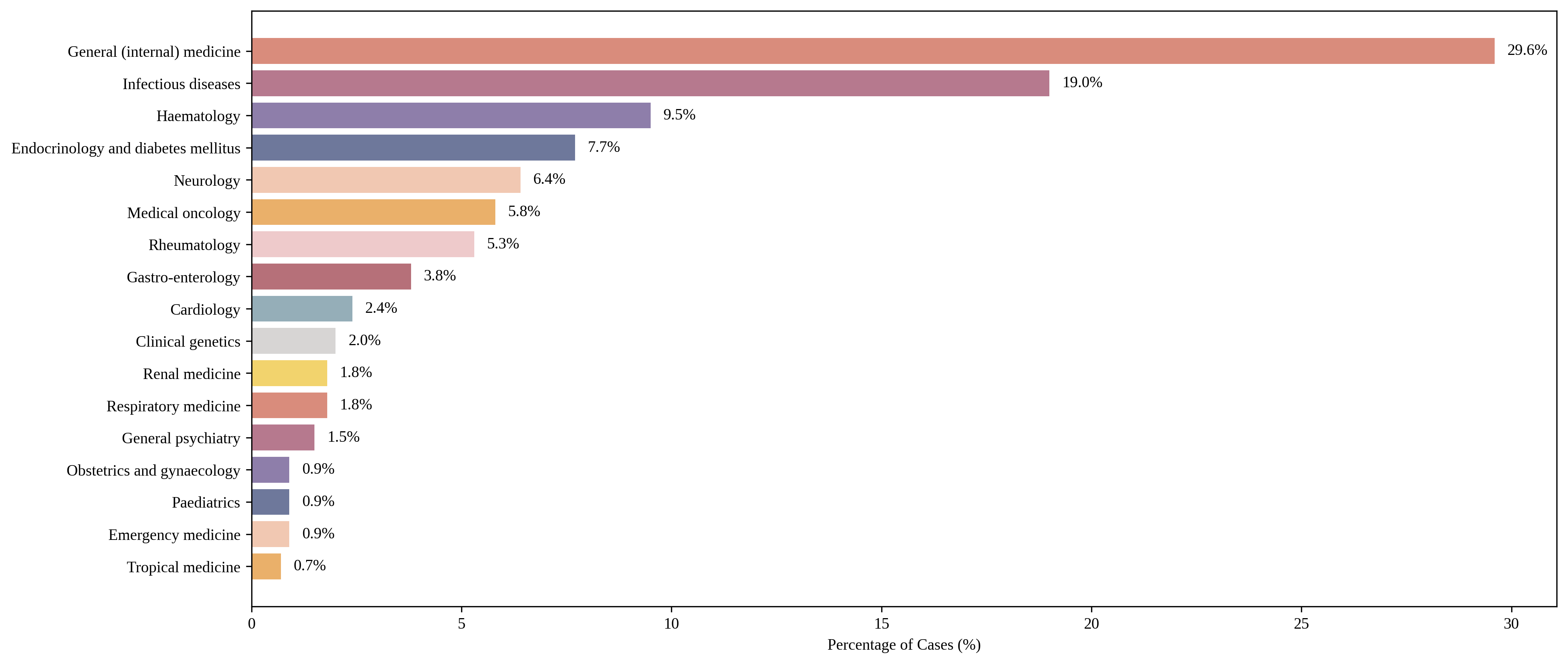}
        \caption{Clinical specialty distribution.}
        \label{fig-spe_dist-supp}
    \end{subfigure}


    \begin{subfigure}[t]{\textwidth}
        \centering
        \includegraphics[width=\textwidth]{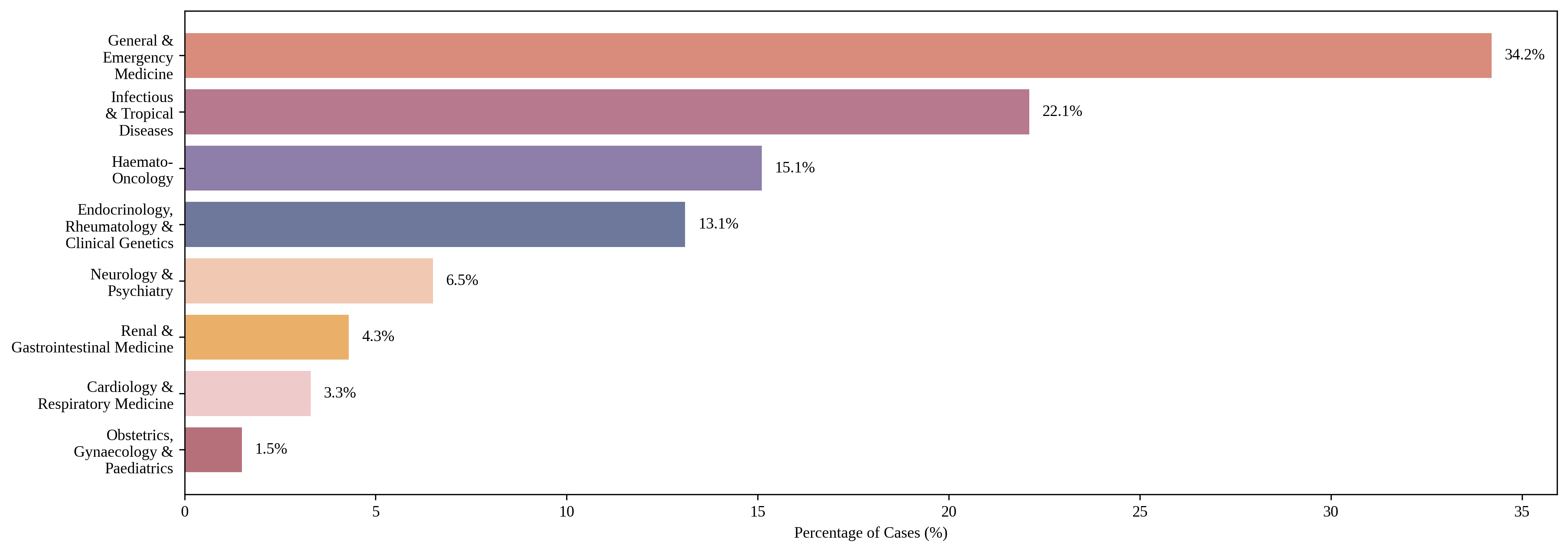}
        \caption{Clinical specialty group distribution.}
        \label{fig-spe_dist_group-supp}
    \end{subfigure}
    
    \caption{Distributions of clinical specialties and specialty groups.}
    \label{fig-supp:sensitivity}
\end{figure*}

We categorize cases into 17 clinical specialties based on UK's General Medical Council Specialties: General (internal) medicine, Infectious diseases, Tropical medicine, Emergency medicine, Haematology, Medical oncology, Endocrinology and diabetes mellitus, Rheumatology, Clinical genetics, Neurology, General psychiatry, Cardiology, Respiratory medicine, Renal medicine, Gastro-enterology, Obstetrics and gynaecology, and Paediatrics. Figure~\ref{fig-spe_dist-supp} shows the distributions of 17 clinical specialties. 

For analysis purposes, we further aggregate these specialties based on expert input to reflect patient management in clinical practice (\emph{e.g.}, similar first-line investigations, inpatient vs. emergency workflows, and typical referral patterns) into 8 groups:
\begin{itemize}
  \item \textbf{General \& Emergency Medicine:} General (internal) medicine; Emergency medicine.
  \item \textbf{Infectious \& Tropical Diseases:} Infectious diseases; Tropical medicine.
  \item \textbf{Haemato-Oncology:} Haematology; Medical oncology.
  \item \textbf{Endocrinology, Rheumatology \& Clinical Genetics:} Endocrinology and diabetes mellitus; Rheumatology; Clinical genetics.
  \item \textbf{Neurology \& Psychiatry:} Neurology; General psychiatry.
  \item \textbf{Cardiology \& Respiratory Medicine:} Cardiology; Respiratory medicine.
  \item \textbf{Renal \& Gastrointestinal Medicine:} Renal medicine; Gastro-enterology.
  \item \textbf{Obstetrics, Gynaecology \& Paediatrics:} Obstetrics and gynaecology; Paediatrics.
\end{itemize}

Figure~\ref{fig-spe_dist_group-supp} demonstrates the distributions of these specialty groups in detail. 

\section{Details of Experiments and Additional Results}
\subsection{Evaluation Metric}
\label{supp-metric}
For open-ended generation, we utilize GPT-5~\cite{openai2025introducinggpt5} for scoring based on the evaluation framework introduced by Kanjee et al.~\cite{kanjee2023accuracy}. The evaluation prompts for DDx and final diagnosis (FDx) are as follows:
\begin{itemize}
    \item Prompt for scoring DDx in open-ended generation:
\begin{lstlisting}[style=appendixprompt]
You are an expert clinician and expert exam-grader. You are grading the quality of the differential diagnosis list for a clinical case.

Inputs
1) Final confirmed diagnosis: {ground_truth}
2) Differential diagnoses: {differential list}

Task
Using medical knowledge, compare the differential diagnoses with the final
diagnosis and assign a quality score on a 0--5 scale.

Scoring rules
5: The actual diagnosis appears explicitly in the differential.
4: A very close diagnosis is suggested (near-miss, closely adjacent).
3: A closely related diagnosis is suggested that could meaningfully help
   workup/management, but is not close in name.
2: A somewhat related diagnosis is suggested but is unlikely to be
   helpful for reaching the target.
0: No suggestions are close or clinically related to the target.

Output format
Score: X
Rationale: <one sentence>

Here are some examples:
...
\end{lstlisting}

\item Prompt for scoring FDx in open-ended generation:
\begin{lstlisting}[style=appendixprompt]
You are an expert clinician and expert exam-grader. You are grading the quality of the final diagnosis for a clinical case.

Inputs
1) Final confirmed diagnosis: {ground_truth}
2) Generated diagnosis: {diagnosis}

Task
Using medical knowledge, compare the generated diagnoses with the final confirmed diagnosis and assign a quality score on a 0--5 scale.

Scoring rules:
5: Exact match to the final diagnosis.
4: Very close diagnosis but not exact.
3: Closely related and plausibly helpful for reaching the final diagnosis
   (same disease family/syndrome; overlapping pathophysiology).
2: Related but unlikely to be clinically helpful (broad category or
   distant relative).
1: Unrelated to the final confirmed diagnosis.

Output format
Score: X
Rationale: <one sentence>

Here are some examples:
...
\end{lstlisting}
\end{itemize}

\subsection{Relative Attention per Token (RAPT)}
\label{supp:rapt}
Relative Attention per Token (RAPT) is defined as the ratio of section-average attention per token to the input-average value in each layer~\cite{liu2025seeing}.
Formally, RAPT is defined as:
\begin{align*}
\text{RAPT}_S^{(l)} & = \frac{\text{Average Attention per Token in } S}{\text{Uniform Attention per Token}} \\
& = \left( \sum_{i \in S} a^{(l)}_i \right) \times \frac{N}{N_S},
\end{align*}
where $N$ is the total number of tokens in the input sequence, $N_S$ is the number of tokens in section $S$, and $a^{(l)}$ is the aggregated attention vector for layer $l$ averaged across all attention heads. For example, a RAPT of 0.1 means each token in that section receives 10\% of the input-average attention.

\subsection{Prompt Refinement}
\label{supp:prompt_refinement}
Prompt after the prompt refinement:
\begin{lstlisting}[style=appendixprompt]
[SYSTEM]
You are an expert clinician tasked with synthesizing complex case files.
Use all provided information to select the most likely diagnosis from the given options.

CRITICAL INSTRUCTION: When pathology images are provided, you MUST prioritize and give PRIMARY WEIGHT to the pathology images in your diagnostic reasoning.
Pathology images are the gold standard for diagnosis and should take precedence over other imaging modalities.
Examine the pathology images carefully and base your diagnosis primarily on the pathological evidence shown in these images.
...
[USER]
...
Remember: Carefully examine and prioritize the pathology images in your diagnostic reasoning.
...
\end{lstlisting}

\subsection{Additional Results}
\label{supp:additional-results}
\begin{figure*}[htbp]
    \centering

    \begin{subfigure}[t]{\textwidth}
        \centering
        \includegraphics[width=\linewidth]{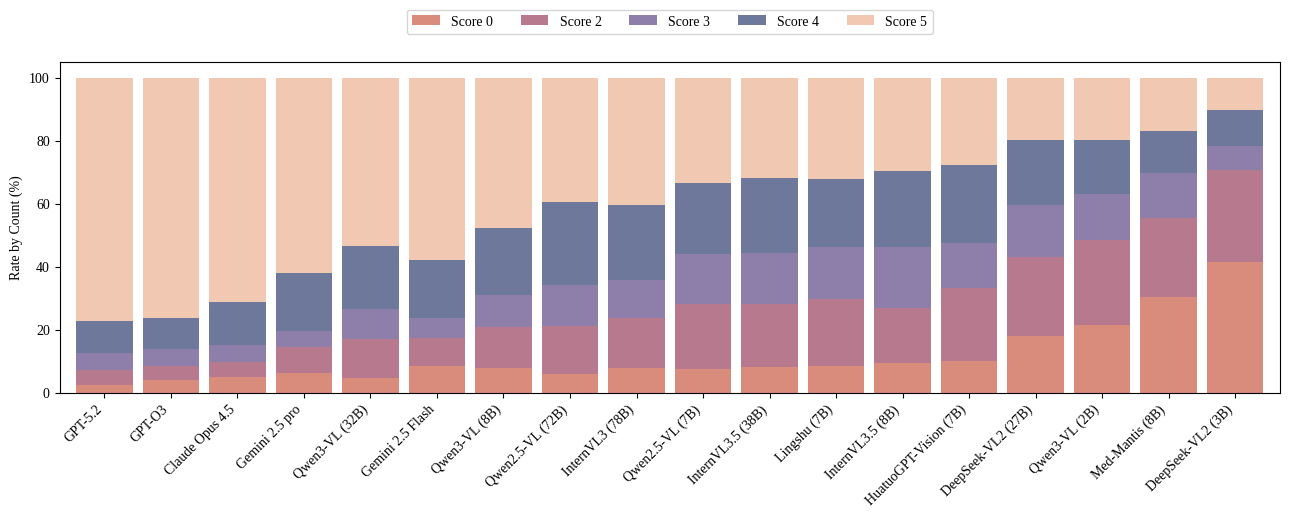}
        \caption{English subset}
        \label{fig-supp:eng_diff_diag_score}
    \end{subfigure}


    \begin{subfigure}[t]{\textwidth}
        \centering
        \includegraphics[width=\linewidth]{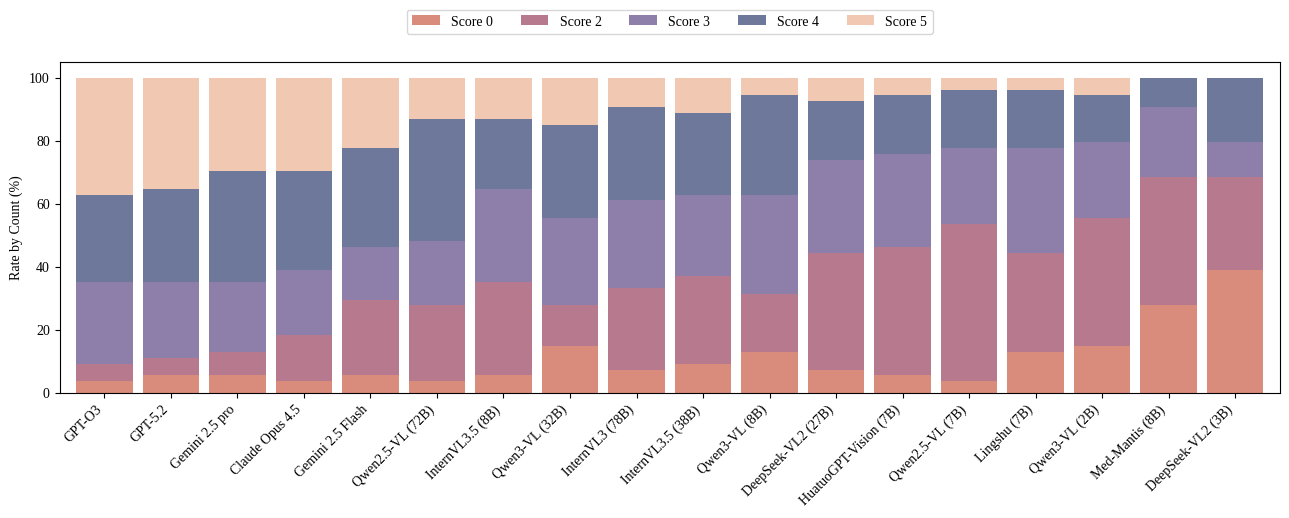}
        \caption{Chinese subset}
    \end{subfigure}
    
    \caption{Distributions of GPT-5 scores on differential diagnosis across models on the English and Chinese subsets. Clinical evidence from diagnostic investigations is provided as raw images.}
    \label{fig-supp:diff-diag}
\end{figure*}

\paragraph{DDx Generation.}
Table~\ref{tab-supp-1} compares average scores of open-ended DDx generation across models. Figure~\ref{fig-supp:diff-diag} visualizes the distribution of scores. We can observe that top-performing models are characterized by a high density of scores of 5, whereas tiny models such as DeepSeek-VL2 (3B) show a higher frequency of scores of 0. We also notice nearly all models exhibit a performance gap between Chinese subset and the English subset, indicated by a significant drop of cases which achieves scores of 5. 

\begin{table}[htbp]
\centering
\setlength{\tabcolsep}{8pt}
\resizebox{\linewidth}{!}{
\begin{tabular}{l | c c c}
\toprule
\textbf{Model} & \multicolumn{3}{c}{\textbf{Average Score $\uparrow$}} \\
\cmidrule(lr){2-4}
 & \textbf{English} & \textbf{Chinese} & \textbf{Overall} \\
\midrule
\textsc{Claude Opus 4.5} & 4.36 & 3.65 & 4.28 \\
\textsc{GPT-O3} & 4.46 & 3.85 & 4.38 \\
\textsc{Gemini 2.5 pro} & 4.15 & 3.70 & 4.10 \\
\textsc{GPT-5.2} & 4.52 & 3.78 & 4.43 \\
\textsc{Gemini 2.5 Flash} & 3.99 & 3.35 & 3.91 \\
\midrule
\textsc{Qwen3-VL (32B)} & 4.00 & 3.02 & 3.88 \\
\textsc{Qwen2.5-VL (72B)} & 3.78 & 3.13 & 3.70 \\
\textsc{InternVL3 (78B)} & 3.70 & 2.96 & 3.61 \\
\textsc{InternVL3.5 (38B)} & 3.42 & 2.85 & 3.35 \\
\textsc{Qwen3-VL (8B)} & 3.35 & 2.85 & 3.29 \\
\textsc{HuatuoGPT-Vision (7B)} & 3.27 & 2.75 & 3.20 \\
\textsc{InternVL3.5 (8B)} & 3.22 & 2.68 & 3.15 \\
\textsc{Lingshu (7B)} & 3.18 & 2.66 & 3.12 \\
\textsc{Qwen2.5-VL (7B)} & 3.05 & 2.49 & 2.98 \\
\textsc{DeepSeek-VL2 (27B)} & 2.81 & 2.74 & 2.80 \\
\textsc{Qwen3-VL (2B)} & 2.65 & 2.41 & 2.62 \\
\textsc{DeepSeek-VL2 (3B)} & 1.78 & 1.74 & 1.78 \\
\textsc{Med-Mantis (8B)} & 2.31 & 1.85 & 2.25 \\
\bottomrule
\end{tabular}}
\caption{Comparison of average GPT-5 scores on differential diagnosis generation across models on the English subset. Clinical evidence from diagnostic investigations is provided as raw images.}
\label{tab-supp-1}
\end{table}

\begin{figure*}[htbp]
    \centering
    \includegraphics[width=\textwidth]{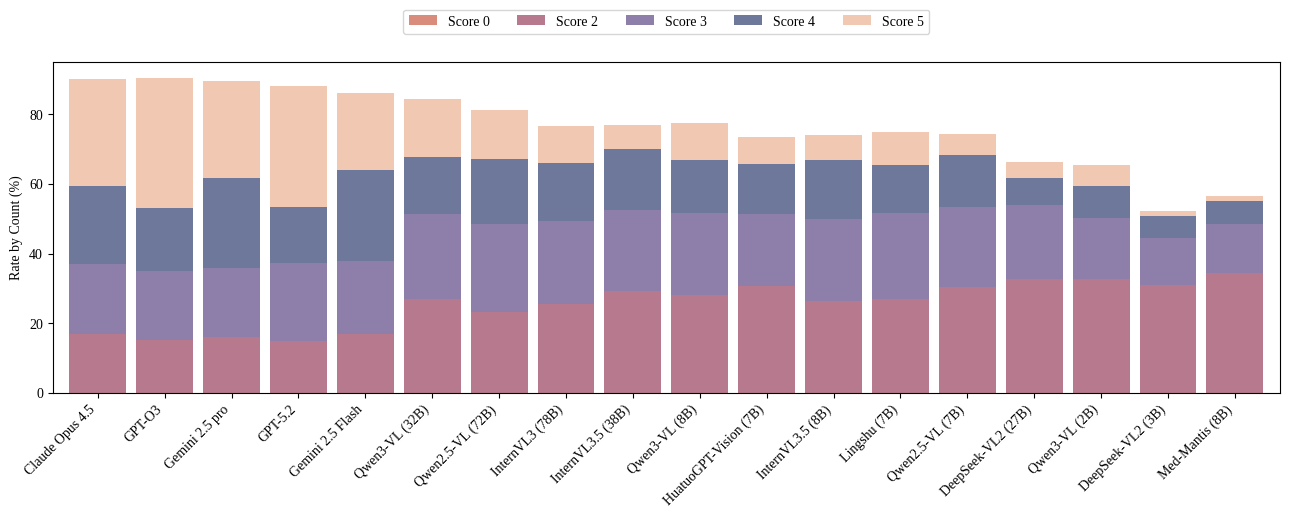}
    \caption{Distributions of GPT-5 scores on open-ended final diagnosis generation across models on the English subset. Clinical evidence from diagnostic investigations is provided as raw images.}
    \label{fig-supp:final-diag}
\end{figure*}

\paragraph{Open-ended FDx Generation.}
In the main paper, we evaluate FDx selection based on MCQs, where models are required to select a FDx from a pre-defined differential list that contains both the correct FDx and distractors. Here, we further evaluate on FDx selection on the English subset of the benchmark by requiring models to output FDx based on their own generated differential lists, termed \emph{open-ended} FDx generation. Outputs are evaluated by GPT-5 against ground-truth FDx (see Appendix~\ref{supp-metric} for metric details). Table~\ref{tab-supp-2} reports the average performance across models. We can observe a similar gap between DDx and FDx performance, where the average score drops $\sim$ 1 on average for leading proprietary models. Figure~\ref{fig-supp:final-diag} details the score distributions. Compared with that of DDx (Figure~\ref{fig-supp:eng_diff_diag_score}), score distribution of FDx (Figure~\ref{fig-supp:final-diag}) exhibits a significant reduction in the proportion of samples with scores $\geq 4$. This is consistent with our previous finding: while models excels at identifying possible medical conditions from heterogeneous clinical evidence (CE) types, they struggle to synthesize them to select the correct FDx. 

\begin{table}[htbp]
\centering
\setlength{\tabcolsep}{8pt}
\resizebox{0.75\linewidth}{!}{
\begin{tabular}{l | c}
\toprule
\textbf{Model} & \textbf{Average Score $\uparrow$} \\
\midrule
\textsc{Claude Opus 4.5} & 3.48 \\
\textsc{GPT-O3} & 3.59 \\
\textsc{Gemini 2.5 pro} & 3.45 \\
\textsc{GPT-5.2} & 3.47 \\
\textsc{Gemini 2.5 Flash} & 3.26 \\
\midrule
\textsc{Qwen3-VL (32B)} & 2.92 \\
\textsc{Qwen2.5-VL (72B)} & 2.86 \\
\textsc{InternVL3 (78B)} & 2.66 \\
\textsc{InternVL3.5 (38B)} & 2.56 \\
\textsc{Qwen3-VL (8B)} & 2.63 \\
\textsc{HuatuoGPT-Vision (7B)} & 2.46 \\
\textsc{InternVL3.5 (8B)} & 2.53 \\
\textsc{Lingshu (7B)} & 2.55 \\
\textsc{Qwen2.5-VL (7B)} & 2.46 \\
\textsc{DeepSeek-VL2 (27B)} & 2.17 \\
\textsc{Qwen3-VL (2B)} & 2.19 \\
\textsc{DeepSeek-VL2 (3B)} & 1.83 \\
\textsc{Med-Mantis (8B)} & 1.88 \\
\bottomrule
\end{tabular}}
\caption{Comparison of average GPT-5 scores on open-ended final diagnosis generation across models on the English subset. Clinical evidence from
diagnostic investigations is provided as raw images.}
\label{tab-supp-2}
\end{table}

\paragraph{Detailed Results on FDx Selection across Clinical Specialties.}
Table~\ref{tab-supp-3} and~\ref{tab-supp-4} detail the accuracy on FDx selection (English subset) across 17 clinical specialties and 8 specialty groups, respectively.

\begin{table*}[htbp]
\centering
\footnotesize
\setlength{\tabcolsep}{2.2pt}
\resizebox{\textwidth}{!}{\begin{tabular}{l | ccccccccccccccccc}
\toprule
\textbf{Model} & 
\makecell[c]{\textbf{Cardio.}} & 
\makecell[c]{\textbf{Clin.}\\ \textbf{Gen.}} & 
\makecell[c]{\textbf{Emerg.}\\ \textbf{Med.}} & 
\makecell[c]{\textbf{Endo.}\\ \textbf{\& Diab.}} & 
\makecell[c]{\textbf{Gastro.}} & 
\makecell[c]{\textbf{Gen.}\\ \textbf{Med.}} & 
\makecell[c]{\textbf{Gen.}\\ \textbf{Psych.}} & 
\makecell[c]{\textbf{Haem.}} & 
\makecell[c]{\textbf{Infect.}\\ \textbf{Dis.}} & 
\makecell[c]{\textbf{Med.}\\ \textbf{Oncol.}} & 
\makecell[c]{\textbf{Neuro.}} & 
\makecell[c]{\textbf{Obs. \&}\\ \textbf{Gyn.}} & 
\makecell[c]{\textbf{Paed.}} & 
\makecell[c]{\textbf{Renal}\\ \textbf{Med.}} & 
\makecell[c]{\textbf{Resp.}\\ \textbf{Med.}} & 
\makecell[c]{\textbf{Rheum.}} & 
\makecell[c]{\textbf{Trop.}\\ \textbf{Med.}} \\
\midrule
\textsc{Claude Opus 4.5} & 71.43 & 44.44 & 100.00 & 68.18 & 69.23 & 65.15 & 71.43 & 64.86 & 64.71 & 52.17 & 84.21 & 0.00 & 33.33 & 50.00 & 83.33 & 57.14 & 66.67 \\
\textsc{GPT-O3} & 42.86 & 55.56 & 50.00 & 68.18 & 69.23 & 65.15 & 71.43 & 70.27 & 65.88 & 56.52 & 63.16 & 33.33 & 33.33 & 25.00 & 66.67 & 61.90 & 100.00 \\
\textsc{Gemini 2.5 Pro} & 42.86 & 55.56 & 75.00 & 77.27 & 69.23 & 57.55 & 42.86 & 67.57 & 67.06 & 65.22 & 63.16 & 66.67 & 66.67 & 50.00 & 83.33 & 52.38 & 100.00 \\
\textsc{GPT-5.2} & 42.86 & 66.67 & 75.00 & 63.64 & 61.54 & 64.39 & 57.14 & 59.46 & 58.33 & 47.83 & 73.68 & 100.00 & 66.67 & 0.00 & 66.67 & 42.86 & 100.00 \\
\textsc{Gemini 2.5 Flash} & 42.86 & 22.22 & 75.00 & 77.27 & 61.54 & 56.06 & 14.29 & 45.95 & 62.35 & 65.22 & 78.95 & 33.33 & 33.33 & 50.00 & 66.67 & 47.62 & 100.00 \\
\midrule
\textsc{Qwen3-VL (32B)} & 42.86 & 55.56 & 75.00 & 50.00 & 61.54 & 49.24 & 42.86 & 51.35 & 44.05 & 65.22 & 68.42 & 33.33 & 0.00 & 0.00 & 50.00 & 47.62 & 66.67 \\
\textsc{Qwen2.5-VL (72B)} & 28.57 & 33.33 & 50.00 & 54.55 & 69.23 & 46.21 & 28.57 & 43.24 & 51.76 & 56.52 & 63.16 & 33.33 & 0.00 & 25.00 & 66.67 & 38.10 & 66.67 \\
\textsc{InternVL3 (78B)} & 42.86 & 44.44 & 50.00 & 45.45 & 53.85 & 45.45 & 14.29 & 40.54 & 40.00 & 52.17 & 68.42 & 33.33 & 33.33 & 25.00 & 50.00 & 42.86 & 66.67 \\
\textsc{InternVL3.5 (38B)} & 42.86 & 0.00 & 50.00 & 59.09 & 38.46 & 43.18 & 14.29 & 51.35 & 38.82 & 52.17 & 57.89 & 100.00 & 0.00 & 25.00 & 33.33 & 42.86 & 66.67 \\
\textsc{Qwen3-VL (8B)} & 28.57 & 33.33 & 25.00 & 59.09 & 53.85 & 37.12 & 28.57 & 45.95 & 40.00 & 47.83 & 52.63 & 33.33 & 0.00 & 0.00 & 16.67 & 38.10 & 33.33 \\
\textsc{HuatuoGPT-V (7B)} & 71.43 & 22.22 & 50.00 & 45.45 & 38.46 & 38.64 & 57.14 & 29.73 & 41.18 & 34.78 & 42.11 & 66.67 & 0.00 & 25.00 & 83.33 & 19.05 & 33.33 \\
\textsc{InternVL3.5 (8B)} & 42.86 & 33.33 & 50.00 & 59.09 & 61.54 & 35.61 & 0.00 & 40.54 & 29.41 & 39.13 & 63.16 & 0.00 & 0.00 & 25.00 & 83.33 & 38.10 & 33.33 \\
\textsc{Lingshu (7B)} & 57.14 & 44.44 & 50.00 & 31.82 & 30.77 & 36.36 & 42.86 & 35.14 & 35.29 & 34.78 & 42.11 & 33.33 & 0.00 & 25.00 & 100.00 & 33.33 & 33.33 \\
\textsc{Qwen2.5-VL (7B)} & 42.86 & 33.33 & 25.00 & 36.36 & 46.15 & 30.30 & 57.14 & 24.32 & 36.47 & 43.48 & 47.37 & 33.33 & 0.00 & 25.00 & 33.33 & 28.57 & 33.33 \\
\textsc{DeepSeek-VL2 (27B)} & 42.86 & 16.67 & 37.50 & 31.82 & 34.62 & 29.92 & 35.71 & 35.14 & 31.76 & 34.78 & 28.95 & 50.00 & 16.67 & 25.00 & 33.33 & 23.81 & 16.67 \\
\textsc{Qwen3-VL (2B)} & 28.57 & 22.22 & 25.00 & 36.36 & 46.15 & 25.00 & 28.57 & 32.43 & 32.94 & 43.48 & 26.32 & 66.67 & 33.33 & 25.00 & 33.33 & 38.10 & 0.00 \\
\textsc{DeepSeek-VL2 (3B)} & 14.29 & 33.33 & 75.00 & 36.36 & 46.15 & 33.33 & 14.29 & 29.73 & 25.88 & 26.09 & 36.84 & 0.00 & 33.33 & 0.00 & 50.00 & 33.33 & 0.00 \\
\textsc{Med-Mantis (8B)} & 57.14 & 22.22 & 50.00 & 31.82 & 30.77 & 25.76 & 28.57 & 27.03 & 28.24 & 34.78 & 36.84 & 0.00 & 0.00 & 25.00 & 16.67 & 14.29 & 66.67 \\
\bottomrule
\end{tabular}}
\caption{Comparison of final diagnosis selection accuracy (\%) across models on the English subset stratified by clinical specialties. Clinical evidence from diagnostic investigations is provided as raw images.}
\label{tab-supp-3}
\end{table*}

\begin{table*}[htbp]
\centering
\small
\setlength{\tabcolsep}{5pt}
\resizebox{\textwidth}{!}{\begin{tabular}{l | c c c c c c c c}
\toprule
\textbf{Model} & 
\makecell[c]{\textbf{Gen. \&} \\ \textbf{Emergency}} & 
\makecell[c]{\textbf{Infectious} \\ \textbf{\& Trop.}} & 
\makecell[c]{\textbf{Haemato-} \\ \textbf{Oncol.}} & 
\makecell[c]{\textbf{Endo, Rheum} \\ \textbf{\& Genetics}} & 
\makecell[c]{\textbf{Neuro. \&} \\ \textbf{Psych.}} & 
\makecell[c]{\textbf{Cardio. \&} \\ \textbf{Resp.}} & 
\makecell[c]{\textbf{Renal \&} \\ \textbf{Gastro.}} & 
\makecell[c]{\textbf{Obst., Gyn.} \\ \textbf{\& Paed.}} \\
\midrule
\textsc{Claude Opus 4.5} & 66.18 & 64.78 & 60.00 & 59.61 & 80.77 & 76.92 & 64.71 & 16.66 \\
\textsc{GPT-O3} & 64.70 & 67.04 & 65.00 & 63.46 & 65.39 & 53.85 & 58.82 & 33.33 \\
\textsc{Gemini 2.5 Pro} & 58.06 & 68.18 & 66.67 & 63.46 & 57.69 & 61.54 & 64.71 & 66.67 \\
\textsc{GPT-5.2} & 64.70 & 59.75 & 55.00 & 55.77 & 69.23 & 53.85 & 47.06 & 83.33 \\
\textsc{Gemini 2.5 Flash} & 56.62 & 63.63 & 53.34 & 55.77 & 61.54 & 53.85 & 58.82 & 33.33 \\
\midrule
\textsc{Qwen3-VL (32B)} & 50.00 & 44.82 & 56.67 & 50.00 & 61.54 & 46.16 & 47.06 & 16.66 \\
\textsc{Qwen2.5-VL (72B)} & 46.32 & 52.27 & 48.33 & 44.23 & 53.85 & 46.15 & 58.82 & 16.66 \\
\textsc{InternVL3 (78B)} & 45.58 & 40.91 & 45.00 & 44.23 & 53.85 & 46.16 & 47.06 & 33.33 \\
\textsc{InternVL3.5 (38B)} & 43.38 & 39.77 & 51.66 & 42.31 & 46.15 & 38.46 & 35.29 & 50.00 \\
\textsc{Qwen3-VL (8B)} & 36.76 & 39.77 & 46.67 & 46.15 & 46.15 & 23.08 & 41.18 & 16.66 \\
\textsc{HuatuoGPT-V (7B)} & 38.97 & 40.91 & 31.67 & 30.77 & 46.16 & 76.92 & 35.29 & 33.34 \\
\textsc{InternVL3.5 (8B)} & 36.03 & 29.54 & 40.00 & 46.15 & 46.16 & 61.54 & 52.94 & 0.00 \\
\textsc{Lingshu (7B)} & 36.76 & 35.22 & 35.00 & 34.61 & 42.31 & 76.92 & 29.41 & 16.66 \\
\textsc{Qwen2.5-VL (7B)} & 30.14 & 36.36 & 31.66 & 32.69 & 50.00 & 38.46 & 41.17 & 16.66 \\
\textsc{DeepSeek-VL2 (27B)} & 30.14 & 31.25 & 35.00 & 25.96 & 30.77 & 38.46 & 32.36 & 33.34 \\
\textsc{Qwen3-VL (2B)} & 25.00 & 31.82 & 36.67 & 34.62 & 26.93 & 30.77 & 41.17 & 50.00 \\
\textsc{DeepSeek-VL2 (3B)} & 34.56 & 25.00 & 28.33 & 34.61 & 30.77 & 30.77 & 35.29 & 16.66 \\
\textsc{Med-Mantis (8B)} & 26.47 & 29.55 & 30.00 & 23.08 & 34.61 & 38.46 & 29.41 & 0.00 \\
\bottomrule
\end{tabular}}
\caption{Comparison of final diagnosis selection accuracy (\%) across models on the English subset stratified by clinical specialty groups. Clinical evidence from diagnostic investigations is provided as raw images.}
\label{tab-supp-4}
\end{table*}

\begin{figure}[t]
    \centering
    \begin{subfigure}[t]{\linewidth}
        \centering
        \includegraphics[width=\linewidth]{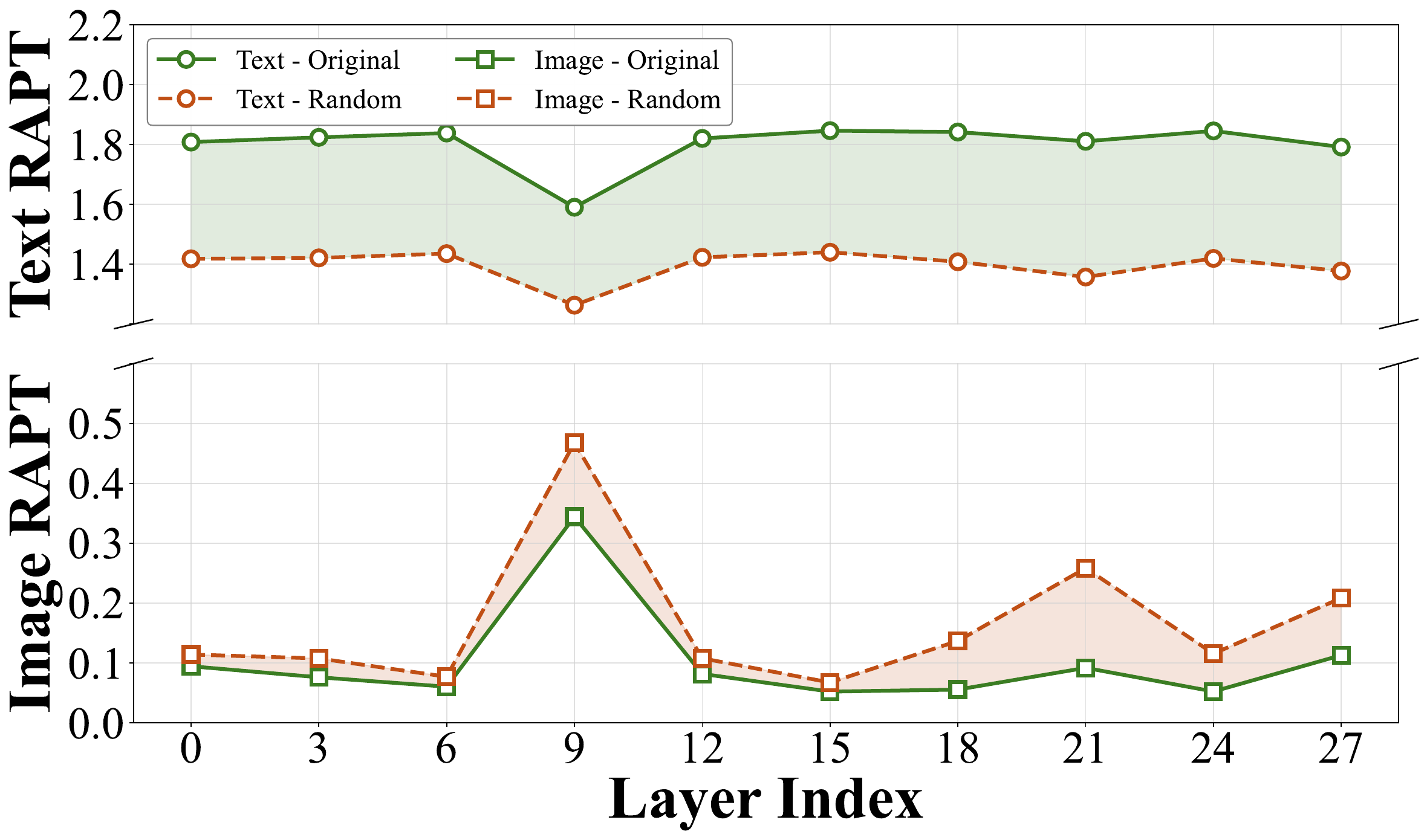}
        \caption{Qwen2.5-VL (7B)}
        \label{fig-supp:rapt-qwen}
    \end{subfigure}


    \begin{subfigure}[t]{\linewidth}
        \centering
        \includegraphics[width=\linewidth]{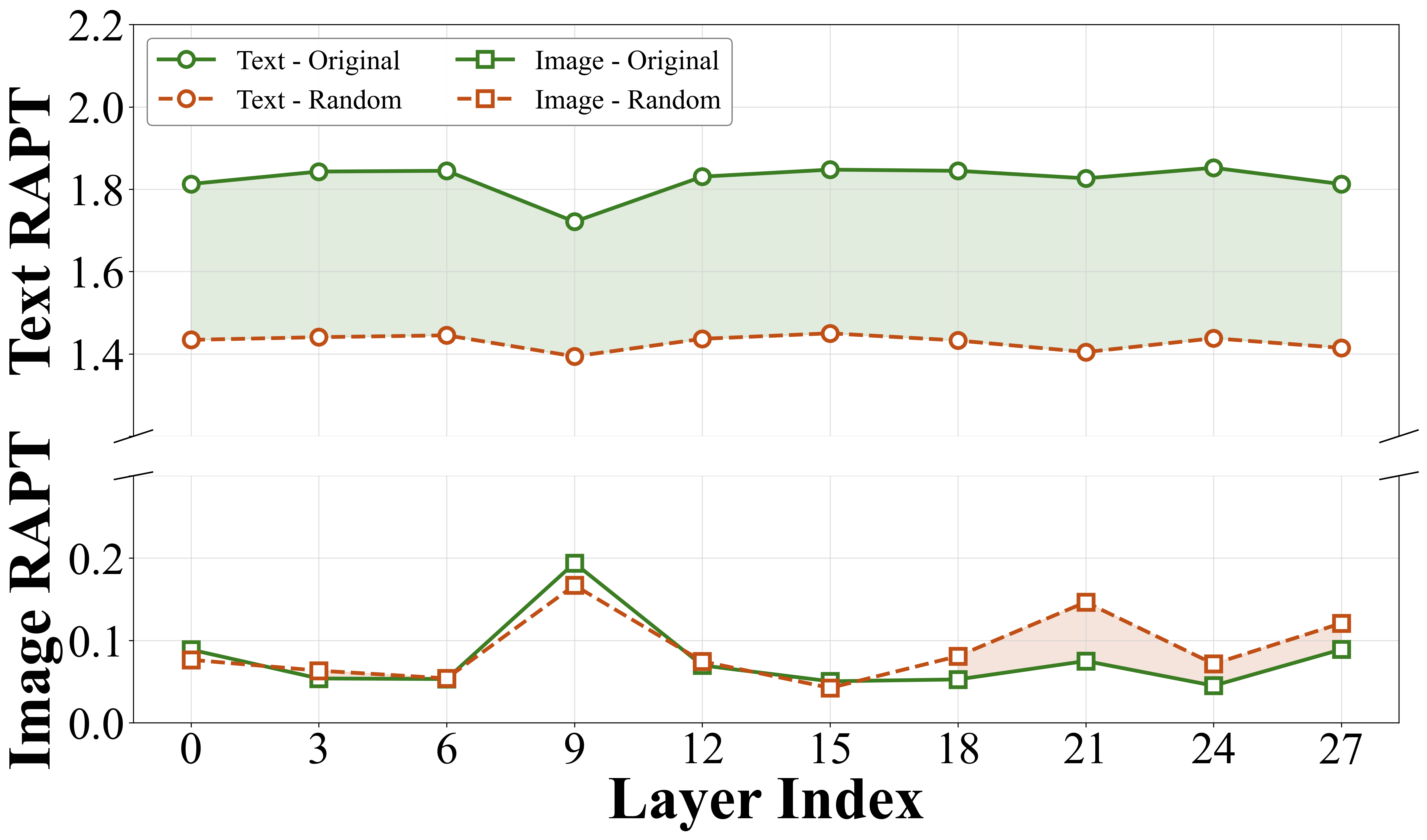}
        \caption{HuatuoGPT-Vision}
        \label{fig-supp:rapt-huatuo}
    \end{subfigure}
    \caption{Layer-wise Relative Attention per Token (RAPT) for Qwen2.5-VL (7B) and HuatuoGPT-Vision on text (excluding question stem) versus image tokens, before and after the \textsc{Random-Text} intervention.}
    \label{fig-supp:rapt}
\end{figure}

\paragraph{Relative Attention per Token (RAPT).}
Figure~\ref{fig-supp:rapt} visualizes the Relative Attention per Token (RAPT) on text (excluding the question stem) and image tokens of Qwen2.5-VL (7B) (Figure~\ref{fig-supp:rapt-qwen}) and HuatuoGPT-Vision (Figure~\ref{fig-supp:rapt-huatuo}) before and after the \textsc{random-text} intervention. We can observe similar trends as Lingshu (Figure~\ref{fig:rapt-lingshu}), discussed in Section~\ref{ana_results} (Finding 1) in the main text. 

\begin{figure*}[htbp]
    \centering

    \begin{subfigure}[t]{0.73\textwidth}
        \centering
        \includegraphics[width=\linewidth]{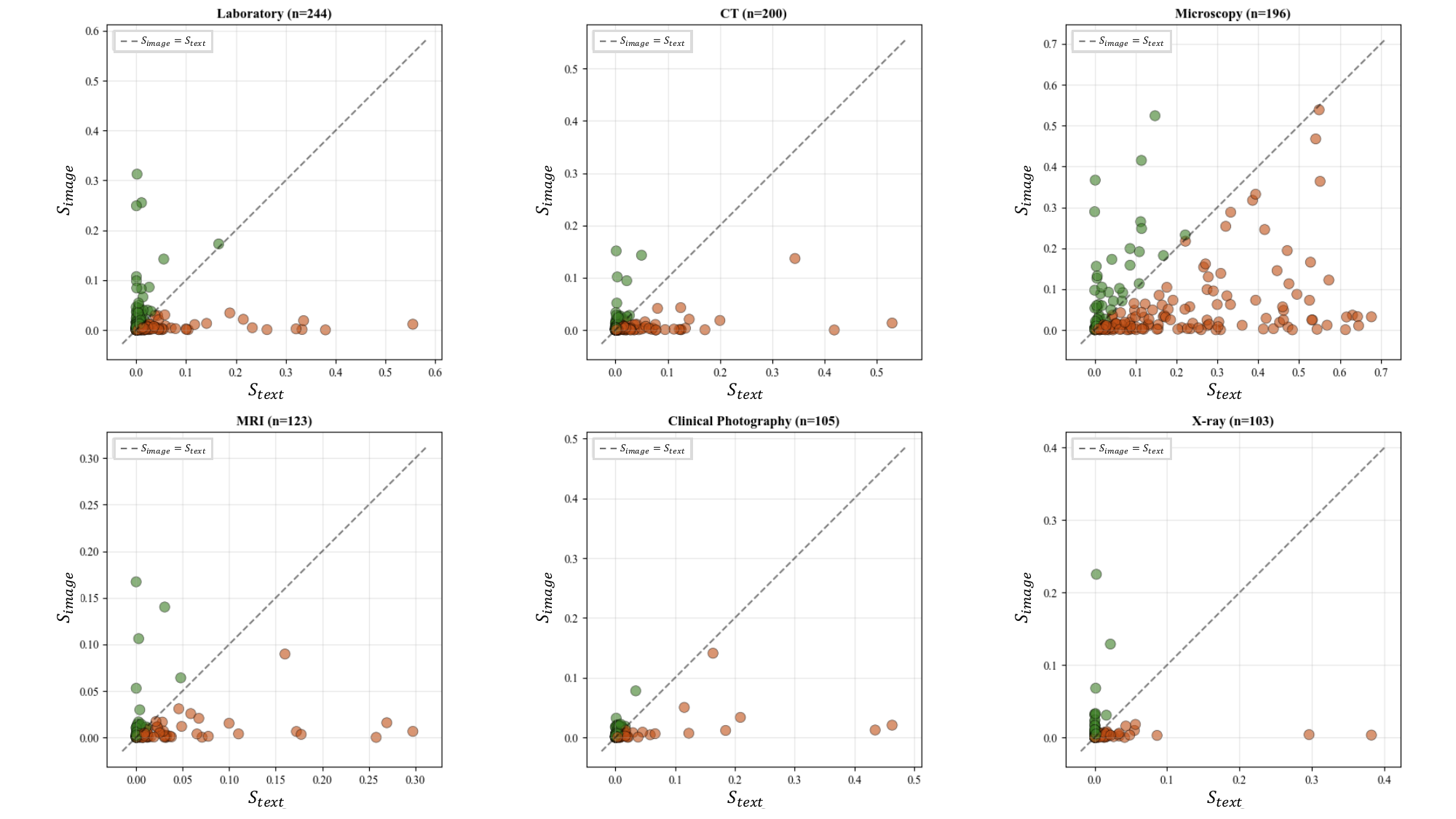}
        \caption{Qwen2.5-VL (7B)}
        \label{fig-supp:qwen-sensitivity}
    \end{subfigure}


    \begin{subfigure}[t]{0.73\textwidth}
        \centering
        \includegraphics[width=\linewidth]{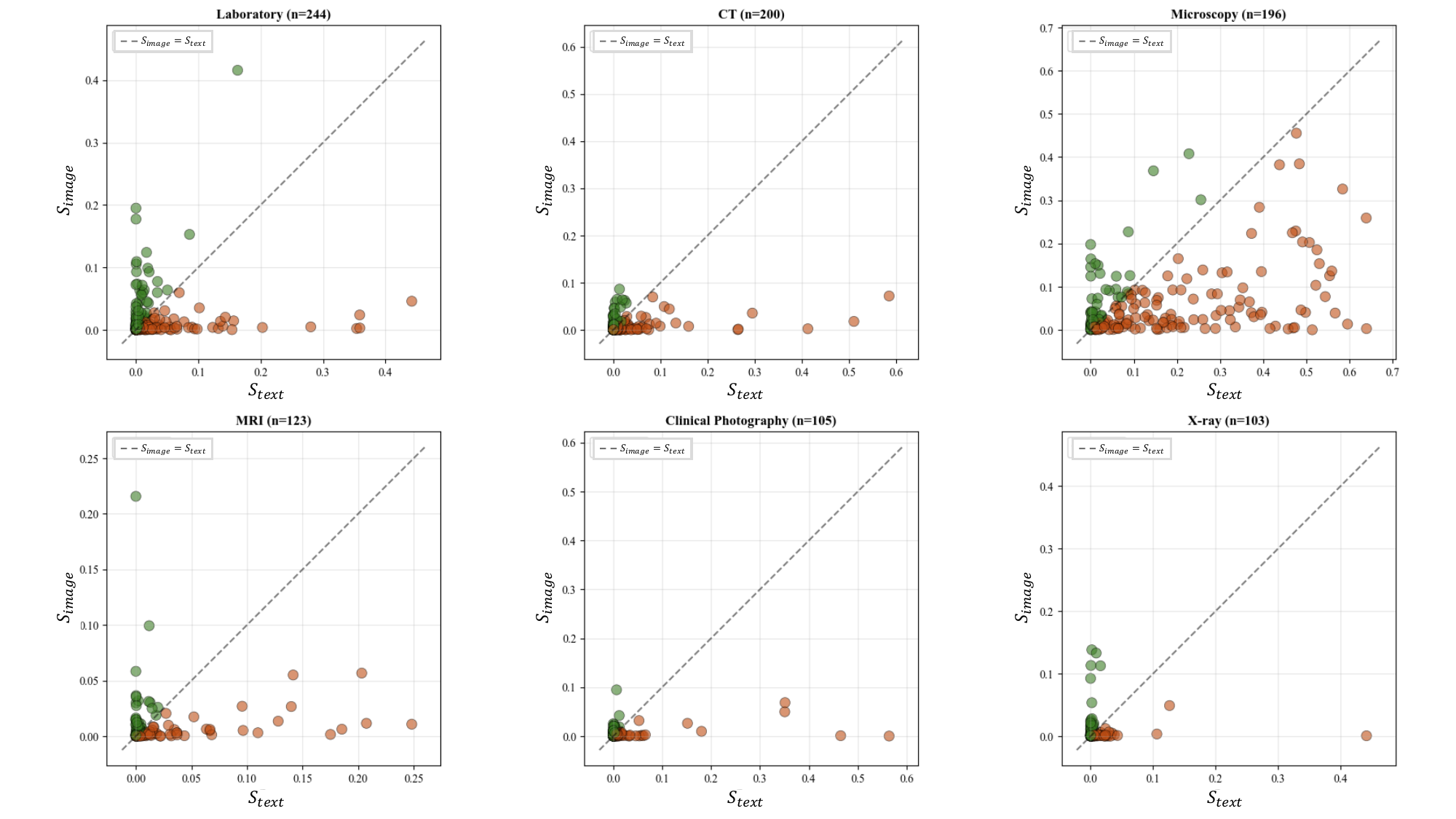}
        \caption{Lingshu (7B)}
        \label{fig-supp:lingshu-sensitivity}
    \end{subfigure}

    \begin{subfigure}[t]{0.73\textwidth}
        \centering
        \includegraphics[width=\linewidth]{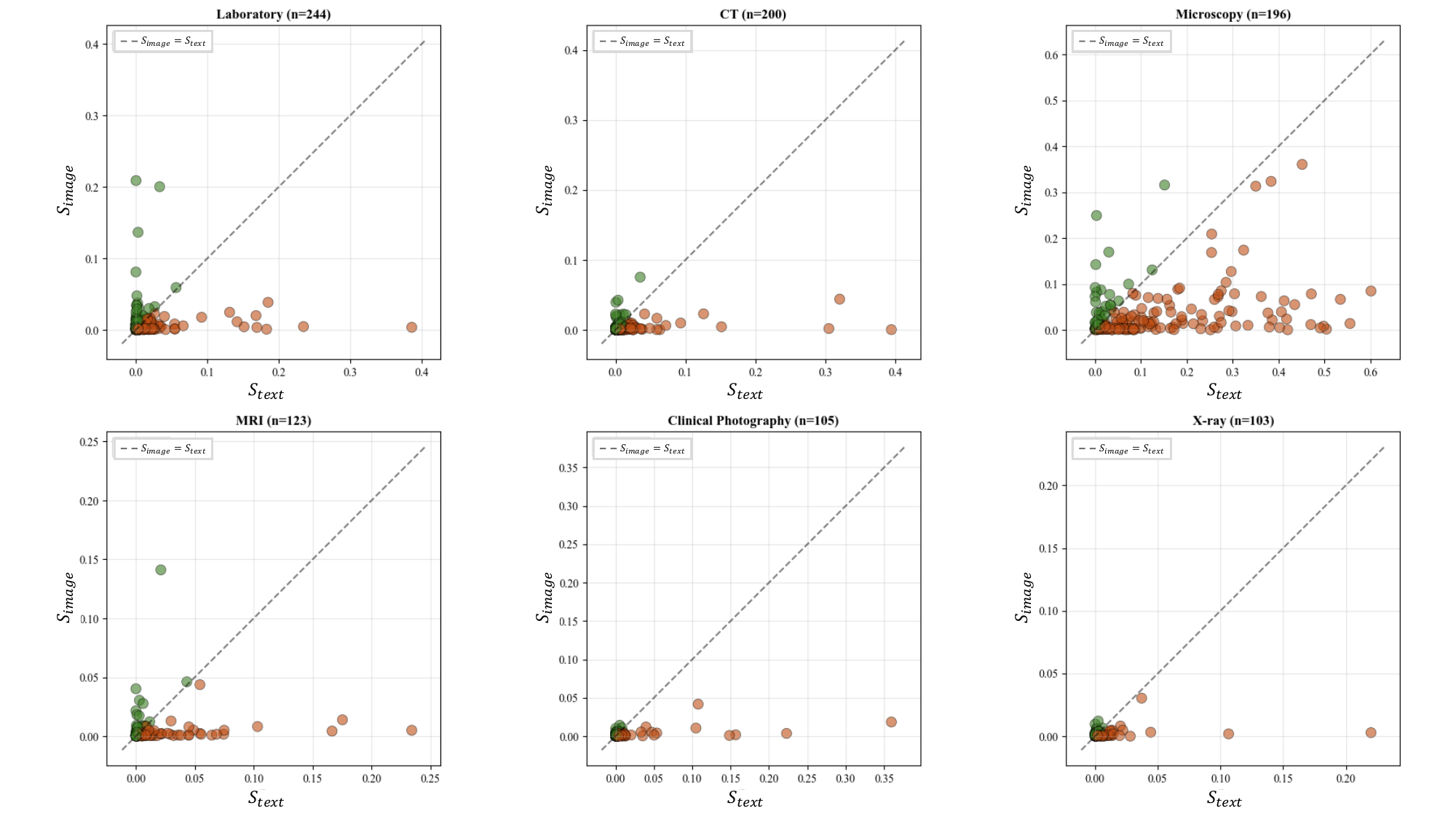}
        \caption{HuatuoGPT-Vision (7B)}
        \label{fig-supp:huatuo-sensitivity}
    \end{subfigure}
    
    \caption{Cross-modal sensitivity across different clinical evidence types. \textcolor{forestgreen}{Green} and \textcolor{burntorange}{brown} points indicate cases where \textcolor{forestgreen}{$S^{(m)}{\text{image}} > S^{(m)}{\text{text}}$} and \textcolor{burntorange}{$S^{(m)}{\text{image}} < S^{(m)}{\text{text}}$}, respectively.}
    \label{fig-supp:sensitivity}
\end{figure*}

\begin{figure*}[htbp]
    \centering

    \begin{subfigure}[t]{0.7\textwidth}
        \centering
        \includegraphics[width=\linewidth]{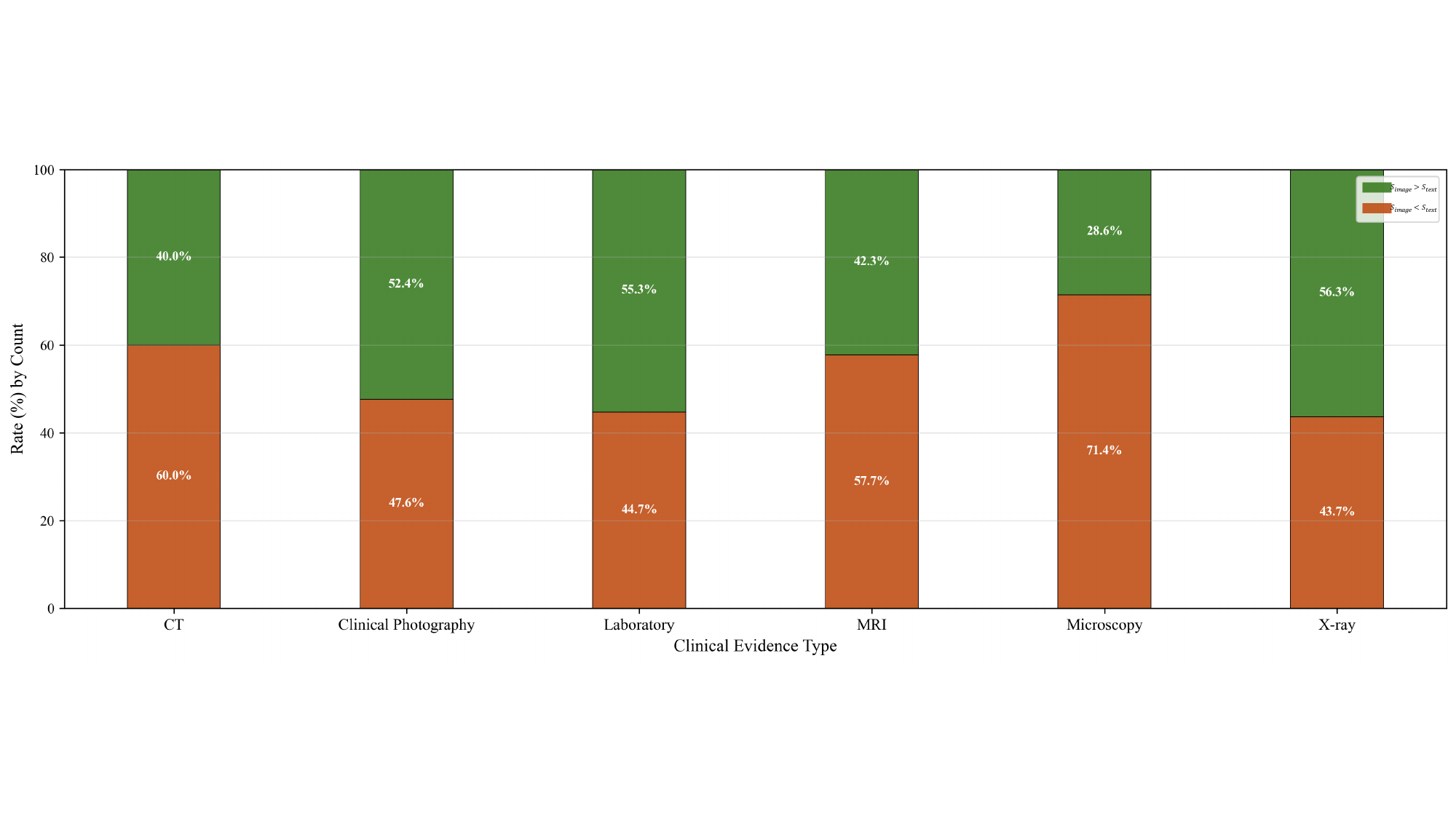}
        \caption{Qwen2.5-VL (7B)}
    \end{subfigure}


    \begin{subfigure}[t]{\textwidth}
        \centering
        \includegraphics[width=0.7\linewidth]{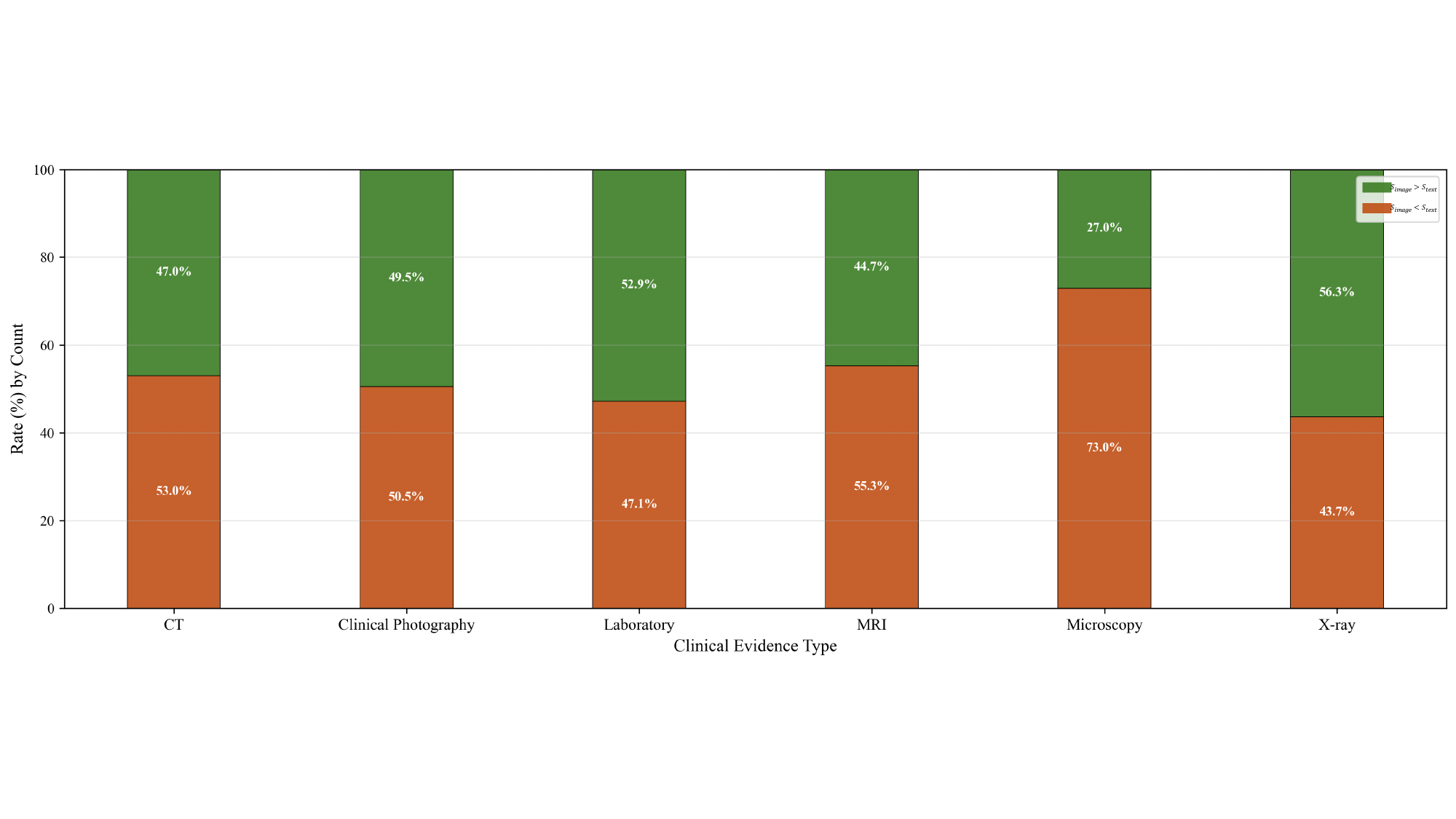}
        \caption{Lingshu (7B)}
    \end{subfigure}

    \begin{subfigure}[t]{\textwidth}
        \centering
        \includegraphics[width=0.7\linewidth]{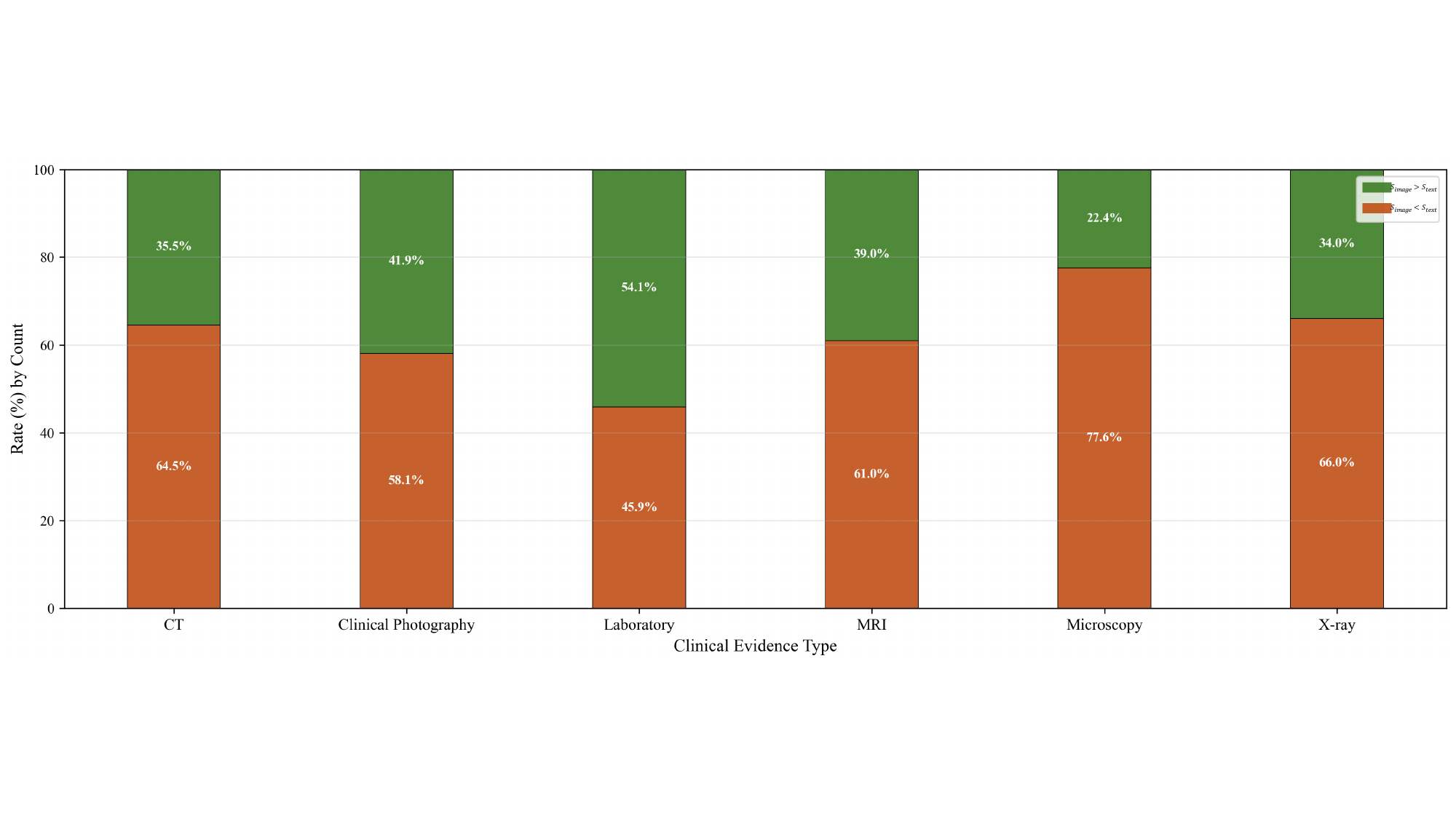}
        \caption{HuatuoGPT-Vision (7B)}
    \end{subfigure}
    
    \caption{Distributions of cases where \textcolor{forestgreen}{$S^{(m)}_{\text{image}}>S^{(m)}_{\text{text}}$ (green)} and \textcolor{burntorange}{$S^{(m)}_{\text{image}}<S^{(m)}_{\text{text}}$ (brown)}.}
    \label{fig-supp:percentage}
\end{figure*}

\paragraph{Cross-modal Sensitivity Comparison.}
Table~\ref{tab-supp-5} shows Normalized Mean Squared Error (NMSE) (Equation~\ref{eq:nmse}) across 6 different clinical evidence (CE) types (\emph{i.e.}, Laboratory, CT, Microscopy, MRI, Clinical Photography and X-ray). Figures~\ref{fig-supp:sensitivity} details cross-modal sensitivity comparison for Qwen2.5-VL (7B) (Figure~\ref{fig-supp:qwen-sensitivity}), Lingshu (Figure~\ref{fig-supp:lingshu-sensitivity}), and HuatuoGPT-Vision (Figure~\ref{fig-supp:huatuo-sensitivity}). We can observe that, beyond CT and Microscopy, the L-shaped distribution persists in Laboratory, MRI, Clinical Photography, and X-ray across all models. These additional results corroborate that the cross-modal utilization gap is a pervasive phenomenon across different CE types. Figures~\ref{fig-supp:percentage} visualizes the distributions of cases where $S^{(m)}_{\text{image}} < S^{(m)}_{\text{text}}$ and $S^{(m)}_{\text{image}} > S^{(m)}_{\text{text}}$. Notably, while Laboratory and X-ray distributions are more balanced, Microscopy remains a distinct outlier: $S^{(m)}_{\text{image}} < S^{(m)}_{\text{text}}$ in over 70\% of cases, peaking at 77.6\% in HuatuoGPT-Vision.

\begin{table}[htbp]
\centering
\setlength{\tabcolsep}{8pt}
\renewcommand{\arraystretch}{1.2}
\resizebox{\linewidth}{!}{\begin{tabular}{lccc}
\toprule
\textbf{Modality} & \textbf{Qwen2.5-VL (7B)} & \textbf{Lingshu} & \textbf{HuatuoGPT-Vision} \\
\midrule
\textsc{Laboratory}        & 1.15 & 1.08 & 1.12 \\
\textsc{CT}                & 0.90 & 0.88 & 0.91 \\
\textsc{Microscopy}         & 0.73 & 0.63 & 0.74 \\
\textsc{MRI}               & 1.00 & 0.99 & 0.99 \\
\textsc{Clinical Photography}          & 0.85 & 0.91 & 0.90 \\
\textsc{X-ray}             & 1.23 & 1.25 & 0.94 \\
\midrule
\textsc{Average}           & 0.98 & 0.95 & 0.93 \\
\bottomrule
\end{tabular}}
\caption{Comparison of normalized mean squared error for different clinical evidence types across models.}
\label{tab-supp-5}
\end{table}

\begin{figure*}[htbp]
    \centering
    \includegraphics[width=\textwidth]{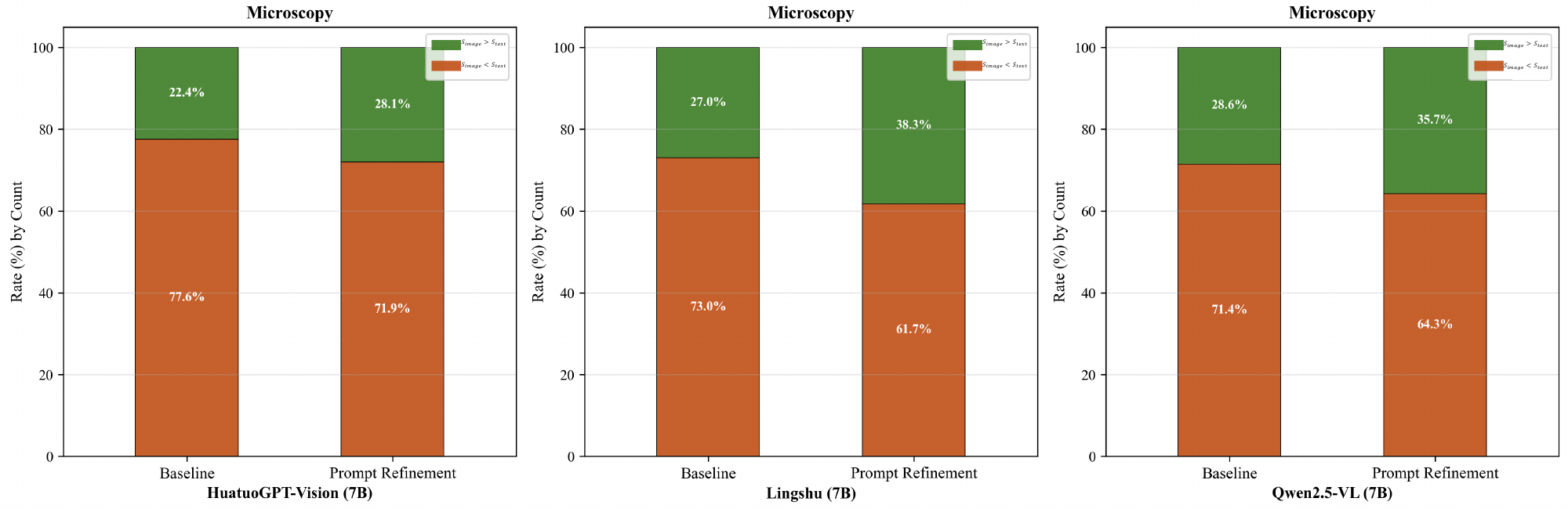}
    \caption{Distributions of cases where \textcolor{forestgreen}{$S^{(m)}_{\text{image}}>S^{(m)}_{\text{text}}$ (green)} and \textcolor{burntorange}{$S^{(m)}_{\text{image}}<S^{(m)}_{\text{text}}$ (brown)} for microscopy before and after prompt refinement.}
    \label{fig-supp:prompt_refinement}
\end{figure*}

\paragraph{Prompt Refinement.}
Table~\ref{tab-supp-6} compares the microscopy NMSE before and after prompt refinement. Figure~\ref{fig-supp:prompt_refinement} visualizes the distributions of cases where $S^{(m)}_{\text{image}} < S^{(m)}_{\text{text}}$ and $S^{(m)}_{\text{image}} > S^{(m)}_{\text{text}}$ for microscopy before and after prompt refinement.

\begin{table}[htbp]
\centering
\setlength{\tabcolsep}{8pt}
\renewcommand{\arraystretch}{1.2}
\resizebox{\linewidth}{!}{\begin{tabular}{lccc}
\toprule
\textbf{Modality} & \textbf{Qwen2.5-VL (7B)} & \textbf{Lingshu} & \textbf{HuatuoGPT-Vision} \\
\midrule
\textsc{Baseline}           & 0.73 & 0.63 & 0.74 \\
\textsc{Prompt Refinement}  &
\cellcolor{red!6}0.67 {(-0.06)} &
\cellcolor{red!2}0.61 {(-0.02)} &
\cellcolor{red!4}0.70 {(-0.04)} \\
\bottomrule
\end{tabular}}
\caption{Comparison of normalized mean squared error (lower is better) for microscopy before and after prompt refinement across models.}
\label{tab-supp-6}
\end{table}

\section{Human Evaluation Details}
\label{supp:human}
\subsection{Physician Demographics}
We recruited two volunteer senior physicians (one male, one female). Both are board-certified in their respective countries of practice and have 35 and 41 years of clinical experience. Participation was voluntary and uncompensated (\emph{i.e.}, no monetary reward). Both participants use English as the language of instruction.

\subsection{Study Procedure}
Our human evaluation study obtained ethical approval, and written consent was collected from each volunteer participating in the assessment. The evaluation was conducted in a local, offline environment without Internet access. Both participants received a standardized briefing describing the study objective, task format, and evaluation interface, followed by a brief training phase (two example cases) to familiarize themselves with the interface and questions. The example cases were not included in the study. For each test case, the interface displays the question as well as multiple pieces of CE (\emph{e.g.}, patient symptoms, laboratory findings, medical imaging scans and microscopy images), and records an answer determined via consensus between both participants to ensure reliability. Upon completing all cases, participants' responses were automatically logged and exported to a CSV file.

\subsection{Evaluation Interface}
\begin{figure*}[htbp]
    \centering
    \includegraphics[width=\textwidth]{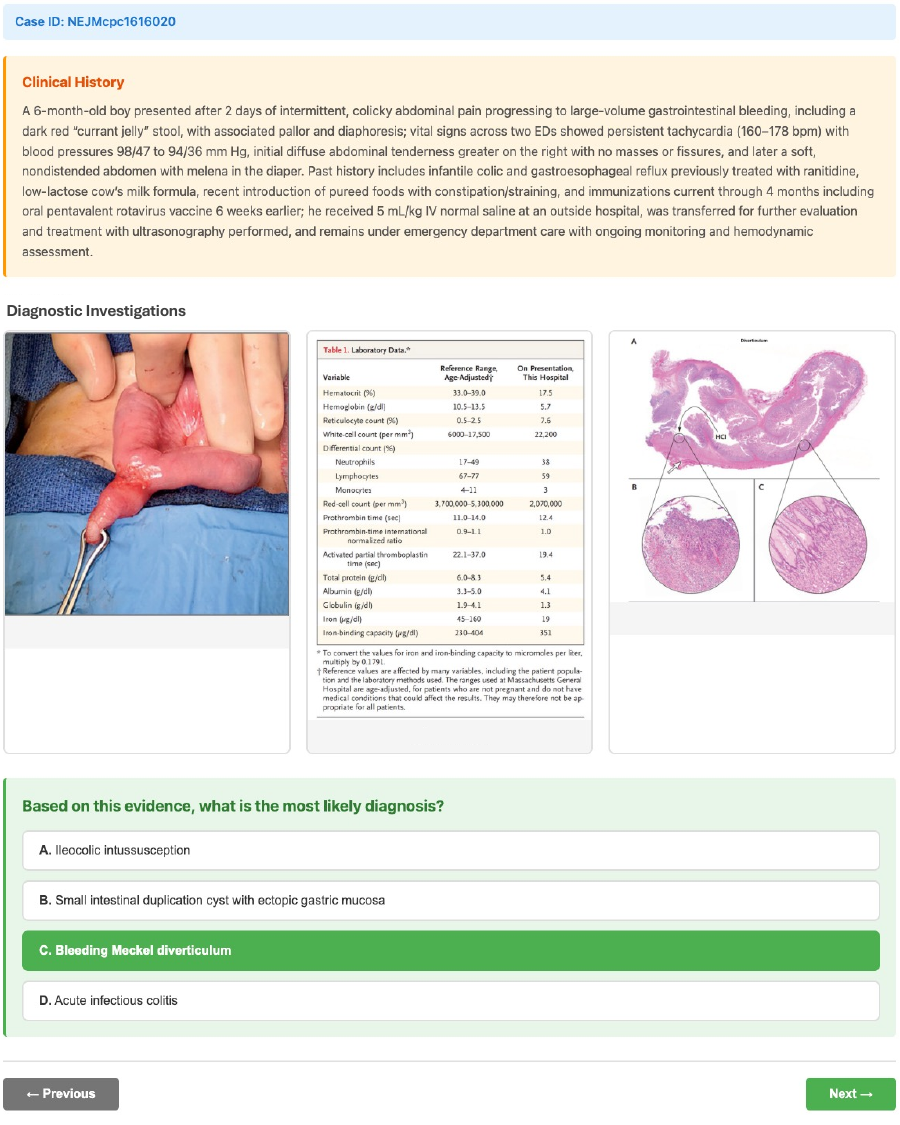}
    \caption{Example of human evaluation interface.}
    \label{fig-supp:human_eval}
\end{figure*}

Figure~\ref{fig-supp:human_eval} illustrates the interface we used to assess expert clinicians’ performance. 

\section{Details on Supervised Fine-tuning}
\label{appendix:sft}
\subsection{Data Construction}
To ensure a fair comparison, we maintained a fixed dataset size of $3{,}000$ samples for both the General and Targeted Supervised Fine-tuning (SFT) experiments.
\begin{itemize}
    \item \textbf{General SFT:} We constructed a baseline dataset by randomly sampling $3{,}000$ instances from the PubMedVision dataset~\cite{chen2024huatuogpt}, representing a general distribution of medical visual instruction tuning tasks following the distribution of the PubMedVision evidence types.
    \item \textbf{Targeted SFT:} We curated a ``microscopy-heavy'' dataset designed to enrich under-utilized modalities. This dataset comprises a balanced mixture of $1{,}500$ microscopy samples from the QUILT-LLaVA Visual Instruct 107K dataset~\cite{quilt-instruct} and $1{,}500$ general medical samples from PubMedVision.
\end{itemize}

\subsection{Training Configuration}
Both the General and Targeted SFT models were trained using an identical configuration to isolate the impact of data composition.
We performed full-parameter fine-tuning on the Qwen2.5-VL (7B)~\cite{yang2025qwen3} model using the AdamW optimizer~\cite{loshchilov2017decoupled} paired with a cosine learning rate scheduler.
The training was conducted on $8 \times$ NVIDIA A6000 GPUs with a global batch size of $128$ and a maximum sequence length of $4{,}096$ tokens.
We utilized a peak learning rate of $5\times10^{-6}$ and trained for $1$ epoch.

\subsection{Additional Results}
We show addition results of targeted SFT on medical domain-specific model (\emph{i.e.}, Lingshu). Table~\ref{tab-supp-7} demonstrates the average accuracy on FDx selection (English subset). We can observe that targeted SFT further improves performance of domain-specific model that has already undergone instruction tuning on medical data.
\begin{table}[htbp]
\centering
\small
\setlength{\tabcolsep}{4.5pt}
\renewcommand{\arraystretch}{1.15}
\begin{tabular}{lc}
\toprule
\textbf{Setting} & \textbf{Acc.} \\
\midrule
\textsc{Baseline} & 36.93 \\
\textsc{Targeted SFT} & \cellcolor{green!3.8}40.70 {(+3.77)} \\
\bottomrule
\end{tabular}
\caption{Final diagnosis accuracy (\%) for Lingshu before and after targeted supervised fine-tuning (SFT).}
\label{tab-supp-7}
\end{table}

\section{Ethical Consideration and Applications}
\subsection{Potential Risks}
Our benchmark is constructed using real clinical cases from the \emph{New England Journal of Medicine Case Record} series and the \emph{National Medical Journal of China} journal. Releasing a dataset derived from these real clinical cases may introduce re-identification risk, particularly for patients with rare conditions or distinctive combinations of findings. Even after privacy safeguards are applied to minimize this risk, residual risk may remain.

\subsection{Data Anonymization Procedures}
The original case records are generally anonymized. Nonetheless, we verify all cases to ensure they were rigorously anonymized. Our verification includes: (i) manual screening for personally identifiable information (PII) and protected health information (PHI); (ii) redacting or generalizing high-risk quasi-identifiers (\emph{e.g.}, uncommon procedures); (iii) removing identifiers embedded in images (\emph{e.g.}, burned-in overlays); and (iv) for clinical photographs, masking patient faces. The released data will exclude physician names and other extraneous identifying details. Where quasi-identifiers remain (\emph{e.g.}, rare combinations of findings), we apply further redaction or generalization prior to release to reduce re-identification risk.

\subsection{Instructions Given To Participants}
\subsubsection{Disclaimer for Annotators}
Thank you for participating in our evaluation. Please read the following before you begin:

\begin{itemize}
    \item \textbf{Voluntary participation:} Your participation is voluntary. You may stop at any time without penalty.
    \item \textbf{Confidentiality:} You will see anonymized materials that exclude personally identifiable information (PII) and protected health information (PHI). Your ratings and your responses will also be kept confidential.
    \item \textbf{Potential discomfort:} Although the task is low risk, some cases may include clinical content (\emph{e.g.}, medical images or descriptions) that could be uncomfortable. You may skip any item or stop at any time.
    \item \textbf{Questions:} If you have questions or concerns during the task, please contact the study organizers.
\end{itemize}

\subsubsection{Instructions for Annotation}
Thank you for participating in our study. Please read the instructions below carefully before you begin.

\paragraph{Task A1: Case Verification and Evidence Completeness.}
For each case, review the clinical history and all associated clinical evidence (text, tables, and figures). Confirm that the evidence is complete and clinically interpretable, and flag any missing, low-quality, or uninterpretable items.

\paragraph{Task A2: Diagnosis Confirmation.}
Verify the final diagnosis reported in the source. If the final diagnosis is not explicitly stated, provide the most supported diagnosis based on the full case report and document the supporting evidence.

\paragraph{Task A3: Evidence--Interpretation Alignment.}
Check that each visual evidence item is correctly paired with its corresponding expert-derived diagnostic interpretation. Flag any misalignment, unsupported interpretation, or statement not grounded in the source.

\subsubsection{Instructions for Experiments}
Thank you for participating in our study. Please read the instructions below carefully.

\paragraph{Task 1: Interface Familiarization}
You will first complete a short training phase with two example cases to familiarize yourself with the interface, the case format, and how clinical evidence is presented. These examples are for practice only and are not included in our evaluation.

\paragraph{Task 2: Differential Diagnosis Generation and Final Diagnosis Selection.}
This task consists of two sub-tasks. Given the case history and all available evidence, you will first provide a ranked differential diagnosis list (from most likely to least likely). Enter all plausible diagnoses in the text box below, separated by commas. Next, click anywhere on the interface to reveal a multiple-choice question for selecting a single final diagnosis. Choose the option most strongly supported by the full set of clinical evidence considered collectively by clicking on your selected option. Your responses are saved automatically.

\paragraph{Task 3: Ambiguity Flags.}
If a case is particularly ambiguous or the evidence is deemed insufficient for a confident decision by you, flag it and provide a brief explanation (\emph{e.g.}, missing key tests, low-quality images or multiple plausible diagnoses).

\subsubsection{Data Consent}
The information you provide in this study will be used only for academic research. Your responses will be stored securely and handled confidentially; we will remove direct identifiers and, where possible, report results only in aggregate to protect your privacy. Participation is voluntary. You may withdraw from the study at any time without penalty, and you may request that your data not be used in our analyses to the extent permitted after withdrawal. If you have any questions about data handling or use, please feel free to contact us.

\subsection{Dataset License}
\begin{itemize}
    \item PubMedVision~\cite{chen2024huatuogpt}: Apache license 2.0
    \item QUILT-LLaVA Visual Instruct 107K~\cite{quilt-instruct}: Creative Commons Attribution Non Commercial No Derivatives 3.0
\end{itemize}

\subsection{Use of AI Assistants in Research}
In our study, generative AI assistants are used sparingly and in accordance with the guidelines on ACL’s Policy on AI Writing Assistance. We utilize ChatGPT for basic paraphrasing and grammar checks. These tools are applied minimally to ensure the authenticity of our work and to adhere strictly to the regulatory standards set by ACL. Our use of these AI tools is focused, responsible, and aimed at supplementing rather than replacing human input and expertise in our research.

\end{document}